\documentclass[letterpaper, 10 pt, journal, twoside]{IEEEtran}  %

\usepackage{graphicx}
\usepackage{subcaption} 

\usepackage{amssymb}
\usepackage{amsmath}
\usepackage{commath}
\usepackage{mathtools}

\usepackage{hyperref}
\usepackage{import}
\usepackage{paralist}
\usepackage{gensymb}
\usepackage{dsfont}
\usepackage{siunitx}
\usepackage[bottom]{footmisc}
\usepackage{framed}
\usepackage{balance} 
\usepackage{color, colortbl}
\usepackage[dvipsnames]{xcolor}
\usepackage[skins]{tcolorbox}
\usepackage{url}
\usepackage{amsthm}
\newtheorem{lemma}{Lemma}
\newtheorem{corollary}{Corollary}
\usepackage{microtype}

\usepackage{amsmath,mathtools}

\usetikzlibrary{shapes,arrows,calc,positioning,fit}

\tikzset{
  block/.style = {draw, rectangle,
    minimum height=1cm,
    align = center
  },
  input/.style = {coordinate,node distance=1cm},
  output/.style = {coordinate,node distance=1cm},
  arrow/.style={draw, -latex,node distance=2cm},
  pinstyle/.style = {pin edge={latex-, black,node distance=2cm}},
  sum/.style = {draw, circle, node distance=1cm},
  gain/.style = {
    regular polygon, regular polygon sides=3,
    draw, fill=white, text width=1em,
    inner sep=0mm, outer sep=0mm,
    shape border rotate=-90
  },
  dot/.style={circle,fill,draw,inner sep=0pt,minimum size=3pt}
}

\newcommand*\diff{\mathop{}\!\mathrm{d}}

\DeclareMathOperator*{\argmin}{arg\,min}
\DeclareMathOperator*{\diag}{\mathrm{diag}}

\newtcolorbox{myframe}[2][]{%
  enhanced,colback=white,colframe=black,coltitle=black,
  sharp corners,boxrule=0.4pt,left=0pt,right=0pt,top=0pt,bottom=0pt,
  fonttitle=\itshape,
  attach boxed title to top left={yshift=-0.3\baselineskip-0.4pt,xshift=2mm},
  boxed title style={tile,size=minimal,left=0.5mm,right=0.5mm,
  colback=white,before upper=\strut},
  title={#2},#1
}

\IEEEoverridecommandlockouts                              %

\title{Modeling of Visco-Elastic Environments \\for Humanoid Robot Motion Control}

\author{Giulio Romualdi$^{1,2}$, Stefano Dafarra$^{1}$, and Daniele Pucci$^{1}$%
\thanks{Manuscript received: October, 15, 2020; Revised December, 17, 2020; Accepted February, 13, 2021.}%
\thanks{This paper was recommended for publication by Editor Kheddar Abderrahmane upon evaluation of the Associate Editor and Reviewers' comments.
This work is supported by An.Dy project which has received funding from the European Union's Horizon 2020 Research and Innovation Programme under grant agreement No. 731540.}
\thanks{$^{1}$Authors are with Dynamic Interaction Control, Istituto Italiano di Tecnologia, Genoa, Italy
        {\tt\footnotesize name.surname@iit.it}}%
\thanks{$^{2}$ First Author is with DIBRIS, University of Genoa, Genoa, Italy}%
}

\begin{document}

\maketitle

\begin{abstract}
 This manuscript presents a model of compliant contacts for time-critical humanoid robot motion control. The proposed model considers the environment as a continuum of spring-damper systems, which allows us to compute the equivalent contact force and torque that the environment
 exerts 
 on the
 contact surface. We show that the proposed model extends the linear and
 rotational springs and dampers -- classically used to characterize soft terrains --
 to the case of \emph{large}  contact surface orientations.
 The  contact model is then used for the 
 real-time whole-body control of humanoid robots walking on visco-elastic environments. The overall approach is validated by simulating
walking motions of 
 the  iCub humanoid robot. Furthermore, the paper compares the proposed whole-body control strategy and state of the art approaches. In this respect, we investigate the terrain compliance that makes the classical approaches  assuming rigid contacts fail. 
 We finally analyze the robustness  of the presented control design with respect to non-parametric uncertainty in the contact-model. 
\end{abstract}
\begin{IEEEkeywords}
Humanoid and Bipedal Locomotion, Whole-Body Motion Planning and Control, Humanoid Robot Systems, Contact Modeling
\end{IEEEkeywords}

\section{Introduction}
\IEEEPARstart{B}{ipedal} locomotion of humanoid robots remains a challenge for the Robotics community despite decades of research on the subject. Among the large variety of complexities that need to be tackled  when addressing the robot locomotion problem, the nature of the terrain is one of the main sources of unpredictable disturbances that impact on robot performances. This paper contributes towards the modeling of soft terrains and the whole-body time-critical control of humanoid robots that locomote 
on compliant floors.

A popular approach to humanoid robot control 
emerged during the DARPA Robotics Challenge. It can be roughly described as 
a hierarchical control architecture composed of several loops \cite{Koolen2016,feng2015optimization}. From outer to inner, one first finds the \emph{trajectory optimization} loop, which generates high-level walking trajectories, e.g., feasible center of mass (CoM) and feet trajectories. The output of this layer feeds the \emph{simplified model control} loop, which is often based on simplified dynamical models such as the Linear Inverted Pendulum Model (LIPM) \cite{Kajita2001} and the Divergent Component of Motion (DCM) \cite{Englsberger2011,Englsberger2013}. 
Then, one has the so called \emph{whole-body quadratic programming (QP) control} loop, whose main objective is to ensure the tracking of the desired trajectories by using  complete robot models. In the whole-body QP control layer, the interaction between the environment and the robot is often modelled using the \emph{rigid contact} assumption~\cite{nava16,doi:10.1142/S0219843619500348,Hopkins2015b,Herzog2016}. Under this hypothesis, the controller can instantaneously change the contact forces to guarantee the tracking of desired quantities. 
If the robot stands on a \emph{visco-elastic} surface, the \emph{rigid contact} assumption no longer holds and the whole-body QP controller cannot arbitrarily change the contact forces.
\par 
At the modeling level, when a robot interacts with a \emph{visco-elastic} surface, it is pivotal to design models that characterize the interaction properties, such as compliance and damping. Remark that the term \emph{contact} describes situations where two bodies come in touch with each other at specific locations~\cite{Gilardi2002}. Then,
 contact models can be classified into two main categories: \emph{rigid} and \emph{compliant}. When the contact is rigid, the bodies' mechanical structure does not change substantially, and the system velocities are often subject to discontinuities~\cite{Whittaker1988}. Although they enable instantaneous feedback controllers for a large variety of robots~\cite{Englsberger2018}, rigid contact models may lead to  poor reproducible  simulation results in the presence of static frictions and multi-body systems \cite{Mason1988a,Stronge1991}. So, the use of \emph{compliant} contact models is a valid alternative to solve the limitations due to the rigid contact formulation. The literature on the non-linear modeling of compliant contacts is extensive, and covering it in details is beyond the scope of this paper~\cite{popov2019handbook,Lankarani1990,Azad2014,Azad2016}.

At the control level, when the contact between a robot and its surrounding  environment is \emph{sufficiently} stiff, controllers based on the \emph{rigid contact} assumption may still lead to \emph{acceptable} robot performances. In this case, 
a possibility is to design contact-model-free passivity-based control strategies~\cite{Henze2016,Mesesan2019}. 
When the contact compliance impairs robot performances, it is pivotal to design contact models that take into account contact stiffness and damping, which can then be incorporated in feedback  controllers.
Pure stiffness (no damping) linear lumped models of the soft contact can, for instance, be considered to design whole-body controllers in the presence of visco-elastic contact surfaces~\cite{10.1115/1.3139652,Flayols2020,Catalano2020}. %
 Damping components can also be added in the contact model~\cite{280780,fahmi2019stance},
but the main assumption remains that the robot makes contact with the environment at isolated points, not on surfaces. 
Still using lumped parameters, another approach to model soft contact surfaces consists in assuming that the contact interaction can be characterized by equivalent springs and rotational dampers, which can then be employed for robot control~\cite{Li2019,doi:10.1177/1729881419897472}. This approach leads to the problem of giving a physical meaning to the contact parameters associated with the torque in the contact wrench. Furthermore, springs and rotational dampers are inherently ill-posed: they make use of the three-angle SO(3) minimal representation to model foot rotations (e.g. the roll-pitch-yaw convention).
At the planning level, the finite element method (FEM) can also be used to model the static equilibrium  of a body in contact with a soft environment~\cite{6100848}. While providing enhanced modeling capabilities, FEM methods demand heavy computational time, which usually forbid their use in time-critical feedback control applications.
\par
This paper contributes towards the modeling of compliant contacts for robot motion control. More precisely, the contributions of this paper follow. $i)$ A new contact model that characterises the stiffness and damping properties of a visco-elastic material that exerts forces and torques on rigid surfaces in contact with it. Differently from classical state of the art models used for robot control,
the presented model considers the environment as a continuum of spring-damper systems, thus allowing us to compute the equivalent contact force and torque by computing surface integrals over the contact surface. As a consequence, we avoid using linear rotational springs and dampers for describing the interaction between robot and environment. We show that the model we propose extends and encompasses these linear models. $ii)$ A whole body controller that allows humanoid robots  walk on visco-elastic floors, which are modeled using the proposed compliant contact model. The robot controller estimates the model parameters on-line, so there is no prior knowledge of the contact parameters. $iii)$ A validation of the approach on the simulated torque-controlled humanoid robot  iCub~\cite{Natale2017}, with an extensive comparison between the proposed methods and classical state of the art techniques.  
Furthermore, we analyze the robustness capability of the presented controller with respect to non-parametric uncertainty in the contact model.
Finally, we present the contact parameter estimation performances in the case of an anisotropic environment.
\par
The paper is organized as follows. Sec.~\ref{sec:background} introduces the notation and recalls some concept of floating base system used for locomotion. Sec.~\ref{sec:contact_model} presents the model used to characterise the interaction between the robot and the environment. Sec~\ref{sec:controller} details the whole-body controller for walking of compliant carpets. Sec~\ref{sec:results} presents the simulation results on the iCub humanoid robot. Sec.~\ref{sec:conclusions} concludes the paper.\looseness=-1 
\section{Background}
\label{sec:background}
\subsection{Notation}
\begin{itemize}
\item $I_n$ and $0_n$ denote the $n \times n$ identity and zero matrices;
\item $\mathcal{I}$ denotes an inertial frame;
\item $\prescript{\mathcal{A}}{}{p}_\mathcal{C}$ is a vector that connects the origin of frame $\mathcal{A}$ and the origin of frame $\mathcal{C}$ expressed in the frame $\mathcal{A}$;
\item given $\prescript{\mathcal{A}}{}{p}_\mathcal{C}$ and $\prescript{\mathcal{B}}{}{p}_\mathcal{C}$,  $\prescript{\mathcal{A}}{}{p}_\mathcal{C} = \prescript{\mathcal{A}}{}{R}_\mathcal{B} \prescript{\mathcal{B}}{}{p}_\mathcal{C} + \prescript{\mathcal{A}}{}{p}_\mathcal{B}= \prescript{\mathcal{A}}{}{H}_\mathcal{B} \begin{bmatrix}
  \prescript{\mathcal{B}}{}{p}_\mathcal{C} ^\top & \;1
\end{bmatrix}^\top$. $\prescript{\mathcal{A}}{}{H}_\mathcal{B}$ is the homogeneous transformations and $\prescript{\mathcal{A}}{}{R}_\mathcal{B} \in SO(3)$ is the rotation matrix; 
\item the \emph{hat operator} is $\hat{.} : \mathbb{R}^3  \to \mathfrak{so}(3)$, where $\mathfrak{so}(3)$ is the set of skew-symmetric matrices and $\hat{x}y = x \times y$. $\times$ is the cross product operator in $\mathbb{R}^3$, in this paper the hat operator is also indicated by $S(.)$;
\item the \emph{vee operator} is $.^\vee : \mathfrak{so}(3) \to \mathbb{R}^3$;
\item $\prescript{\mathcal{A}}{}{{v}}_\mathcal{B} \in \mathbb{R}^3$ denotes the time derivative of the relative position between the frame $\mathcal{B}$ and the frame $\mathcal{A}$;
\item $\prescript{\mathcal{A}}{}{\omega}_\mathcal{B} \in \mathbb{R}^3$ denotes the angular velocity between the frame $\mathcal{B}$ and the frame $\mathcal{A}$ expressed in the frame $\mathcal{A}$;
\item The velocity of a frame $\mathcal{B}$ w.r.t. the frame $\mathcal{A}$ is uniquely identified by the twist $\prescript{}{}{\mathrm{v}}_\mathcal{B}^ \top = \begin{bmatrix} \prescript{\mathcal{A}}{}{{v}}_\mathcal{B} ^ \top & \prescript{\mathcal{A}}{}{\omega}_\mathcal{B}^ \top \end{bmatrix}$.
\end{itemize}

\subsection{Humanoid Robot Model}
A humanoid robot is modelled as a floating base multi-body system composed of $n+1$ links connected by $n$ joints with one degree of freedom each. Since none of the robot links has an a priori pose w.r.t. the inertial frame $\mathcal{I}$, the robot configuration is completely defined by considering both the joint positions $s$ and the homogeneous transformation from the inertial frame to the robot frame (i.e. called base frame $\mathcal{B}$). 
In details, the configuration of the robot can be uniquely determined by the triplet $q = (\prescript{\mathcal{I}}{}{p}_\mathcal{B}, \prescript{\mathcal{I}}{}{R}_\mathcal{B}, s) \in  \mathbb{R}^3 \times SO(3) \times \mathbb{R}^n$.
The velocity of the floating system is represented by the triplet $ \nu = (\prescript{\mathcal{I}}{}{v}_\mathcal{B}, \prescript{\mathcal{I}}{}{\omega}_\mathcal{B}, \dot{s})$, where $\prescript{\mathcal{I}}{}{v}_\mathcal{B}$ and $\dot{s}$ are the time derivative of the position of the base and the joint positions, respectively. $\prescript{\mathcal{I}}{}{\omega}_\mathcal{B}$ is defined as $\prescript{\mathcal{I}}{}{\omega}_\mathcal{B} = ( \prescript{\mathcal{I}}{}{\dot{R}}_\mathcal{B} \prescript{\mathcal{I}}{}{R}_\mathcal{B} ^\top)^\vee$.

\par
Given a frame attached to a link of the floating base system, its position and orientation w.r.t. the inertial frame is uniquely identified by a homogeneous transformation, $\prescript{\mathcal{I}}{}{H}_\mathcal{A} \in SE(3)$.
The map from $\nu$ to its twist $\prescript{}{}{\mathrm{v}}_\mathcal{A}$ is linear and its matrix representation is the well known Jacobian matrix $J_\mathcal{A}(q)$: $\prescript{}{}{\mathrm{v}}_\mathcal{A} = J_\mathcal{A}(q) \nu$.
For a floating base system the Jacobian can be split into two sub-matrices. One multiplies the base velocity while the other the joint velocities. The frame acceleration is $\prescript{}{}{\dot{\mathrm{v}}}_\mathcal{A} = J_\mathcal{A}(q) \dot{\nu} + \dot{J}_\mathcal{A}(q) \nu$.
The dynamics of the floating base system can be described by the Euler-Poincar\'e equation~\cite{Marsden2010}:
\begin{equation}
\label{eq:robot_dynamics}
    M(q) \dot{\nu} + h(q, \nu) = B \tau + \sum_{k = 1}^{n_c} J^\top_{\mathcal{C}_k}(q) f_k,
\end{equation}
where, $M(q) \in \mathbb{R} ^{(n + 6) \times(n + 6)}$ is the mass matrix, $h(q, \nu) \in \mathbb{R} ^{n + 6}$ is contains the Coriolis, the centrifugal and the gravitational terms.  $B$ is a selector matrix, while $\tau \in \mathbb{R}^n$ are the joint torques. $f_k \in \mathbb{R}^6$ is the contact wrench expressed in a frame located on the body in contact, and oriented as the inertial frame.
$n_c$ indicates the number of contact wrenches. Henceforth, we assume that at least one of the link is in contact with the environment, i.e. $n_c \ge 1$.
\par
The dynamics~\eqref{eq:robot_dynamics} can be expressed by separating the first $6$ rows, referring to the non-actuated floating base, from the last $n$ rows referring to the actuated joints as:
\begin{IEEEeqnarray}{L}
\IEEEyesnumber \phantomsection
M_{\nu}(q,\nu) \dot{\nu} + h_{\nu} (q, \nu) = \sum_{k = 1}^{n_c} J^\top_{{\mathcal{C}_k}_\nu}(q)  f_k, \IEEEyessubnumber \label{eq:robot_dynamics_base}\\
M_{s}(q,\nu) \dot{\nu} + h_{s} (q, \nu) = \sum_{k = 1}^{n_c} J^\top_{{\mathcal{C}_k}_s}(q)  f_k + \tau. \IEEEyessubnumber \label{eq:robot_dynamics_joints}
\end{IEEEeqnarray}
The centroidal momentum of the robot ${}_G h$ is the aggregate linear and angular momentum of each link of the robot referred to the robot center of mass (CoM). 
It is worth recalling that the rate of change of the centroidal momentum of the robot can be expressed by using the external contact wrenches acting on the system~\cite{nava16}
\begin{equation}
\label{eq:centroidal_momentum_dynamics}
    {}_G \dot{h} = \sum_{k = 1}^{n_c}\begin{bmatrix}
      I_3 & 0_3 \\
      S(p_{\mathcal{C}_k} - p_{\mathcal{C}}) & I_3 
    \end{bmatrix} f_k + mg.
\end{equation}
\section{Modelling of Visco-Elastic Environments  \label{sec:contact_model}}

\begin{figure}[t]
\centering
\includegraphics[width=1\columnwidth]{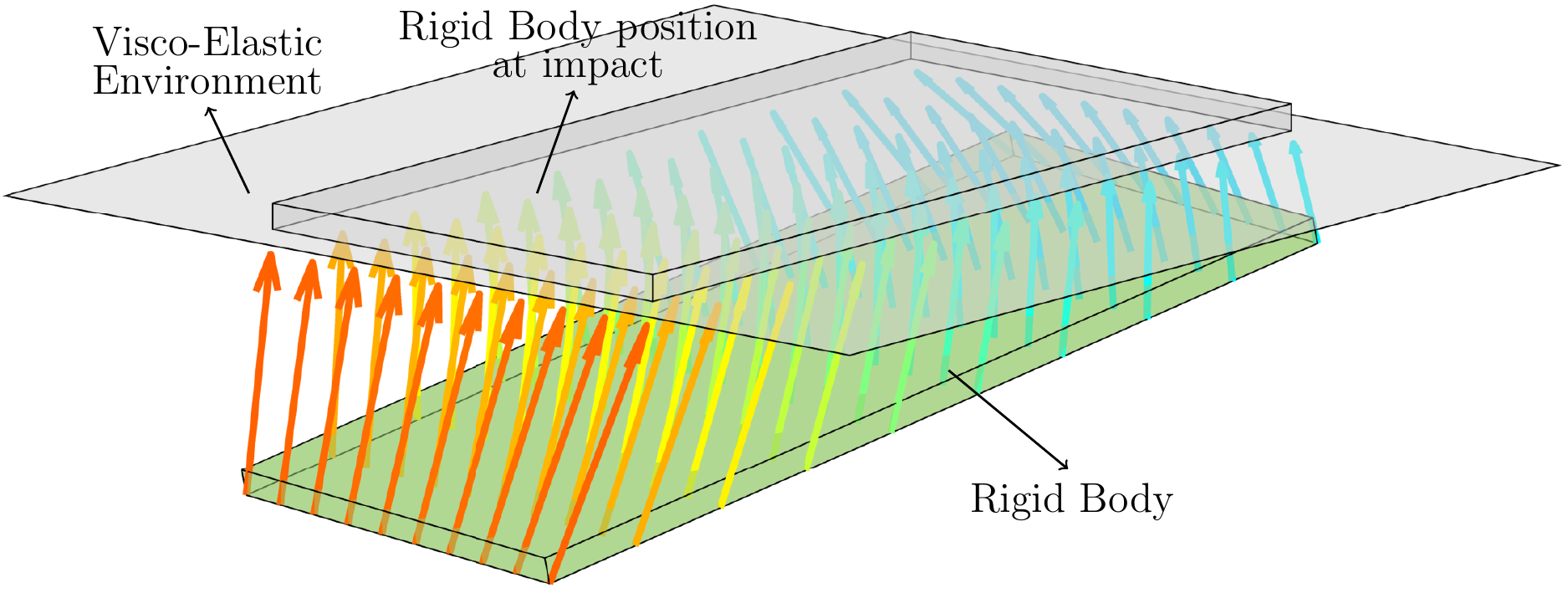}
\caption{Vector field generated by~\eqref{eq:contact_model_general}.\label{fig:contact_model_vector_field}
}
\end{figure}

Consider a rigid body that makes a contact with a visco-elastic surface, and assume that:
\begin{enumerate}
    \item the environment characteristics are isotropic;
    \item there exists a contact domain $\Omega \subset \mathbb{R}^3$;
    \item  $\forall x \in \Omega$, there exists a continuous pure force distribution that depends on the point $x$ and its velocity $\dot{x}$, i.e.
    \begin{equation}
        \rho : \mathbb{R}^3 \times \mathbb{R}^3  \rightarrow \mathbb{R}^3.
    \end{equation}
\end{enumerate}
Given the above assumptions, the contact torque distribution about a point $x_0 \in \mathbb{R}^3$, $\mu_{x_0} : \mathbb{R}^3 \times \mathbb{R}^3  \rightarrow \mathbb{R}^3$ writes
    \begin{equation}
        \mu_{x_0}(x, \dot{x}) = S(x - x _ 0) \rho(x, \dot{x}).
    \end{equation}
Once the pure force and torque distributions are defined, then the equivalent contact wrench $f = [f_l; f_a]$ is given by the following integration over the contact domain $\Omega$, i.e.
    \begin{equation}
    \label{eq:contact_wrench_generic}
        f = \begin{bmatrix}
        \int \int _ {\Omega} \rho(x, \dot{x}) \diff \Omega \\
         \int \int _\Omega  \mu_{x_0}(x, \dot{x}) \diff\Omega \\
        \end{bmatrix}.
    \end{equation}
Next Lemma proposes a model that can be used to describe a contact between a body and a compliant environment.
\begin{lemma}
\label{lemma:compliant_model}
Let $\bar{\mathcal{X}}$ the set of the points $\bar{x}\in\mathbb{R}^3$:
\begin{equation}
     \bar{\mathcal{X}} = \{ \bar{x}\in\mathbb{R}^3  : \rho(\bar{x}, 0) = 0  \}.
\end{equation}
Assume that: $i)$ the contact domain $\Omega$ is a rectangle of dimensions $l$ and $w$; $ii)$ the point $x_0 {\in} \mathbb{R}^3$ is the center of the rectangular domain;  $iii)$ the distribution $\rho$ is given by:
\begin{equation}
\rho(x, \dot{x}) = k ( \bar{x} - x) - b  \dot{x}, \quad k >0,\; b>0.
\label{eq:contact_model_general}
\end{equation}
\par
Then, the equivalent contact force and torque are given by
\begin{IEEEeqnarray}{LL}
\IEEEyesnumber
\label{eq:contact_force_integral_rectangle}
f _ l =& lw |e _ 3^\top \prescript{\mathcal{I}}{}R _\mathcal{F} e _ 3| [k (\bar{p} - p) - b \dot{p}] \\
\label{eq:contact_torque_integral_rectangle}
f_a =&  \frac{l w}{12} |e _ 3^\top \prescript{\mathcal{I}}{}{R} _\mathcal{F} e _ 3| \nonumber \\
&\left\{l^2  S(\prescript{\mathcal{I}}{}{R} _\mathcal{F} e_1)  [b  S(\prescript{\mathcal{I}}{}{R} _\mathcal{F} e_1) {}^{\mathcal{I}} \omega _ {\mathcal{F}} + k   \prescript{\mathcal{I}}{}{\bar{R}} _\mathcal{F} e_1]  \right .  \\
& + \left . w^2  S(\prescript{\mathcal{I}}{}{R} _\mathcal{F} e_2)  [b  S(\prescript{\mathcal{I}}{}{R} _\mathcal{F} e_2) {}^{\mathcal{I}} \omega _ {\mathcal{F}} + k   \prescript{\mathcal{I}}{}{\bar{R}} _\mathcal{F} e_2] \right\} \nonumber,
\end{IEEEeqnarray}
where $\prescript{\mathcal{I}}{}R _\mathcal{F}$ is the rotation from the inertial frame $\mathcal{I}$ to a frame rigidly attached to the body $\mathcal{F}$.
$p$ is the origin of $\mathcal{F}$ w.r.t. $\mathcal{I}$. $\dot{p}$ and ${}^{\mathcal{I}} \omega _ {\mathcal{F}}$ are the linear and angular velocity of the rigid body. $\bar{p}$ and  $\prescript{\mathcal{I}}{}{\bar{R}} _\mathcal{F}$ are the position and the rotation of the frame $\mathcal{F}$ such that the wrench is equal to zero. 
\end{lemma}

The proof is in the Appendix. Lemma~\ref{lemma:compliant_model} shows that \eqref{eq:contact_wrench_generic} depends only on the contact force distribution $\rho$ and the shape of the contact domain $\Omega$. As a consequence, we avoid using rotational springs and dampers for describing the interaction between robot and environment. Furthermore, Lemma~\ref{lemma:compliant_model} also contains a close solution for the equivalent contact wrench \eqref{eq:contact_force_integral_rectangle}  \eqref{eq:contact_torque_integral_rectangle} in case of a linear contact model~\eqref{eq:contact_model_general} and a rectangular contact surface $\Omega$.
To give the reader a better comprehension of the Lemma~\ref{lemma:compliant_model}, we can imagine that the set $\bar{\mathcal{X}}$ contains the position of all the points of the foot sole at the touch down.  
Figure~\ref{fig:contact_model_vector_field} shows the vector field generated by \eqref{eq:contact_model_general} in case of a rectangular contact surface. Finally, it is important to recall that since the contacts are unilateral, this model is valid as long as the normal forces are positive and the tangential component lies inside the friction cone.

The  model~\eqref{eq:contact_force_integral_rectangle}-\eqref{eq:contact_torque_integral_rectangle} also encompasses the classical linear modeling techniques for soft terrains. The corollary below shows, in fact, that linear approximations of~\eqref{eq:contact_force_integral_rectangle}-\eqref{eq:contact_torque_integral_rectangle} lead to  linear and rotational springs and dampers usually used to model contact wrenches due to soft terrains~\cite{doi:10.1177/1729881419897472}.
\begin{corollary}
\label{corollary:approximation}
Let $f_l$ and $f_a$ the  contact force and torque given by \eqref{eq:contact_force_integral_rectangle} and \eqref{eq:contact_torque_integral_rectangle}, respectively. Assume that $\prescript{\mathcal{I}}{}{\bar{R}} _\mathcal{F} = I _3 $ and  $\prescript{\mathcal{I}}{}{R} _\mathcal{F}$ is approximated with its first order of the Taylor expansion, i.e. $\prescript{\mathcal{I}}{}R _\mathcal{F} = I + S(\Theta)$, with $\Theta \in \mathbb{R}^3$. Assume that $\Theta$ represents the classical sequence of roll-pitch-yaw, namely $\prescript{\mathcal{I}}{}R _\mathcal{F}(\Theta) = R_z(\Theta_3)R_y(\Theta_2)R_x(\Theta_1)$.
\\
Then, the contact model \eqref{eq:contact_force_integral_rectangle}-\eqref{eq:contact_torque_integral_rectangle} writes 
\begin{equation}
\label{eq:linear_model}
    f_l^l = \mathcal{K}_l (\bar{p} - p) - \mathcal{B}_l \dot{p}, \quad f_a^l = - \mathcal{K}_a \Theta - \mathcal{B}_a \dot{\Theta},
\end{equation}
with 
$\mathcal{K}_l = l w k I_3$, $\mathcal{K}_a =  k l w / 12 \diag(\begin{bmatrix} w ^ 2 & l ^ 2 & l ^2 {+} w ^2 \end{bmatrix})$,   $\mathcal{B}_l = l w b I _3$, and  $\mathcal{B}_a =  b l w / 12 \diag(\begin{bmatrix} w ^ 2 & l ^ 2 & l ^2 {+} w ^2 \end{bmatrix})$.
\end{corollary}
The proof of Corollary~\ref{corollary:approximation} is obtained by substituting $\prescript{\mathcal{I}}{}R _\mathcal{F} = I + S(\Theta)$ into \eqref{eq:contact_force_integral_rectangle} and \eqref{eq:contact_torque_integral_rectangle}.  
Corollary~\ref{corollary:approximation} thus shows that
the model~\eqref{eq:contact_force_integral_rectangle}-\eqref{eq:contact_torque_integral_rectangle} 
extends the 
linear models~\cite{doi:10.1177/1729881419897472} to the case of \emph{large} contact surface orientations. 
Note, in fact, that the linear model~\eqref{eq:linear_model}, as well as the analogous in~\cite[Eq.(8)]{doi:10.1177/1729881419897472}, is ill-posed for \emph{large} orientations because of the   singularities of the parametrization $\Theta$.

\section{Whole-body Controller \label{sec:controller}}

The whole-body controller aims to track kinematic and dynamic quantities.
The proposed whole-body controller computes the desired joint torques using the robot joint dynamics~\eqref{eq:robot_dynamics_joints}, where the robot acceleration $\dot{\nu}$ is set to the desired \emph{starred} quantity and the contact wrenches $f_k$ are estimated/measured.
The desired generalized robot acceleration $\dot{\nu}^{*}$ is chosen to track the desired centroidal momentum trajectory, the torso and root orientation, and the feet pose, while considering the contact model presented in Section~\ref{sec:contact_model}.
\par 
The control problem is formulated using the stack of tasks approach. The control objective is achieved by framing the controller as a constrained optimization problem where the low priority tasks are embedded in the cost function, while the high priority tasks are treated as constraints.

\subsection{Low and high priority tasks \label{sec:tasks}}
What follows presents the tasks required to evaluate the desired generalized robot acceleration $\dot{\nu}^{*}$. 
\subsubsection{Centroidal momentum task}
In case of visco-elastic contacts, the contact wrench $f_k$ cannot be arbitrarily chosen. From now on, we assume we can control the contact wrench derivative $\dot{f}_k$. Thus, by differentiating the centroidal system dynamics \eqref{eq:centroidal_momentum_dynamics}, we obtain:
\begin{equation}
    \label{eq:centroidal_momentum_derivative}
    {}_G \ddot{h} = \sum_{k = 1}^{n_c} \begin{bmatrix}
  0_3 & 0_3 \\
  S(v_{\mathcal{C}_k} - v_\mathcal{C}) & 0_3
  \end{bmatrix} f_k + 
  \begin{bmatrix}
      I_3 & 0_3 \\
      S(p_{\mathcal{C}_k} - p_{\mathcal{C}}) & I_3 
    \end{bmatrix}
    \dot{f}_k.
\end{equation}
In order to follow the desired centroidal momentum trajectory, we minimize the weighted norm of the error between the robot centroidal momentum and the desired trajectory:
\begin{equation}
    \Psi_h = \frac{1}{2} \norm{{}_G \ddot{h} ^ * -{}_G \ddot{h}},
\end{equation}
where $\Lambda _ h$ is a positive definite diagonal matrix. $f_k$ is the estimated/measured contact wrench. ${}_G\ddot{h}^*$ is the desired centroidal momentum derivative and it is in charge of stabilizing the desired centroidal momentum dynamics:

\begin{equation}
\begin{split}
    {}_G\ddot{h}^* &= {}_G\ddot{h}^{ref} + k_h^d ({}_G\dot{h}^{ref} - {}_G\dot{h}) + k_h^p ({}_G h^{ref} - {}_G h) \\ 
    &+ k_h^i \int {}_G h^{ref} - {}_G h \diff t,
    \end{split}
\end{equation}
where the integral of the angular momentum is computed numerically.
Here, $k^d_h$, $k^p_h$, and $k^i_h$ are three diagonal matrices. When the equality holds, the centroidal dynamics converges exponentially to the reference value if and only if $k^p_h$, $k^i_h$ and $k_h^d k_h^p -k_h^i$ are positive definite.  

\subsubsection{Torso and root orientation tasks}
While walking, we require the torso and the root robot frames to have a specific orientation w.r.t. the inertial frame. To accomplish this task, we minimize the norm of the error between a desired angular acceleration and the actual frame angular acceleration:
\begin{equation}
    \Psi_\circ = \frac{1}{2} \norm{\dot{\omega}_\circ ^ {*} - (\dot{J} _{\circ} \nu + J _{\circ} \dot{\nu})}^2  _ {\Lambda _ \circ},
\end{equation}
where the subscript $\circ$ indicates the root $\mathcal{R}$ and torso $\mathcal{T}$ frames. $\Lambda _ \circ$ is a positive definite matrix that weighs the contributions in different directions. $\dot{\omega} ^*_\circ$ is set to guarantee almost global stability and convergence of $\prescript{\mathcal{I}}{}{R}_{\circ}$ to $ \prescript{\mathcal{I}}{}{R}_{\circ} ^{ref}$ \cite{Olfati-Saber:2001:NCU:935467}:

\begin{IEEEeqnarray}{LL}
\IEEEyesnumber
\dot{\omega}^{*}_\circ &= \dot{\omega}^{ref} - c_0 \left(\hat{\omega} \prescript{\mathcal{I}}{}{R}_{\circ} \prescript{\mathcal{I}}{}{R}_{\circ} ^{ref ^ \top} - \prescript{\mathcal{I}}{}{R}_{\circ} \prescript{\mathcal{I}}{}{R}_{\circ} ^{ref^{\top}} \hat{\omega}^{ref}\right)^{\vee} \nonumber\\
&- c_1 \left(\omega - \omega^{ref}\right) - c_2 \left(\prescript{\mathcal{I}}{}{R}_{\circ} \prescript{\mathcal{I}}{}{R}_{\circ} ^{{ref}^{\top}}\right) ^\vee. \label{eq:rotational_pid_acceleration}
\end{IEEEeqnarray}

Here, $c_0$, $c_1$, and $c_2$ are positive numbers.

\subsubsection{Swing foot task}
Concerning the tracking of the swing foot trajectory, we minimize the following cost function
\begin{equation}
    \Psi_{\mathcal{F}} = \frac{1}{2} \norm{\dot{\mathrm{v}}_{\mathcal{F}} ^ {*} - (\dot{J} _{\mathcal{F}} \nu + J _{\mathcal{F}} \dot{\nu})}^2 _ {\Lambda _ {\mathcal{F}}},
\end{equation}
the angular part of $\dot{\mathrm{v}}^{*}_{\mathcal{F}}$ is given by \eqref{eq:rotational_pid_acceleration} where the subscript $\circ$ is substitute with $\mathcal{F}$, while the linear part $\dot{v}^{*}$ is equal to $\dot{v}^{*}_{\mathcal{F}} =  \dot{v}\,^{ref}_{\mathcal{F}} - k^d _{x _{f}} (v_{\mathcal{F}} - v^{ref}_{\mathcal{F}}) - k^p _{x _{f}} (p_{\mathcal{F}} - p^{ref}_{\mathcal{F}})$.
Here, the gains are again positive definite matrices.

\subsubsection{Regularization tasks}
In order to prevent the controller from providing solutions with huge joint variations, we introduce a regularization task for the joint variables. The task is achieved by asking for a desired joint acceleration that depends on the error between the desired and measured joint values, namely:
\begin{equation}
    \Psi_s = \frac{1}{2} \norm{k_s^p(s^{ref} -s) - k_s^d \dot{s} - \ddot{s}}^2  _ {\Lambda _ s},
\end{equation}
where $s^{ref}$ is a desired joint configuration, $k_s^d$, $k_s^p$ and $\Lambda _ s$ are symmetric positive definite matrices. 
To reduce the amount of the contact wrench required to accomplished the centroidal momentum tracking, the following task is considered:
\begin{equation}
    \Psi_{f_k} = \frac{1}{2} \norm{k_f^p(f_k^{ref} -f_k) - \dot{f}_k}^2  _ {\Lambda _ {f_k}}.
\end{equation}
Here, $\Lambda _ {f_k}$ and $k_f^p$ are positive definite matrices. $f_k$ is the estimated/measured contact wrench and $f^{ref}_k$ is the desired force regularization value.
\subsubsection{Contact Wrench feasibility}
The feasibility of the contact wrenches $f_k$ is in general guaranteed via another set of inequalities of the form:
\begin{equation}
    \label{eq:contact_wrench_feasibility_constraint}
    A f_k \le b.
\end{equation}
More specifically, $f_k$ has to belong to the associated friction cone, while the position of the local CoP is constrained within the support polygon. However, in  the case of visco-elastic contacts, the contact wrenches cannot be arbitrarily chosen. This limitation can be overcome by discretizing the contact wrench dynamics using the forward Euler method:
\begin{equation}
    \label{eq:contact_force_discerization}
    f_k^{i+1} = f_k^{i} + \dot{f}_k^{i} T,
\end{equation}
where $T$ is the constant integration time. 
We can require the contact wrench at the next instant to guarantee the inequality constraints \eqref{eq:contact_wrench_feasibility_constraint}.
Combining \eqref{eq:contact_force_discerization} and \eqref{eq:contact_wrench_feasibility_constraint}, we can now obtain a tractable set of inequality constraints: $A \dot{f}_k T \le b -  A f_k$, 
where the superscript $i$ has been dropped, and $f_k$ represents the measured/estimated contact wrench.
\subsubsection{Contact model dynamics}
The contact wrench dynamics can be obtained by differentiating equations \eqref{eq:contact_force_integral_rectangle} and \eqref{eq:contact_torque_integral_rectangle}. 
By computing the time derivative of $f$, one has $\dot{f}_k = h_{f_k} + g_{f_k} \dot{\mathrm{v}}_{\mathcal{F}_k}$.
where $h_{f_k} \in \mathbb{R}^6$ and $g_{f_k}\in\mathbb{R}^{6 \times 6}$ is full rank for each possible admissible state.
\subsection{Optimal Control Problem}
The control objective is achieved by casting the control problem as a constrained optimization problem whose conditional variables are $\dot{\nu}$ and $\dot{f}_k$ where $k$ represents the foot in contact with the environment, namely left, right or both.
Since the tasks presented in Sec~\ref{sec:tasks} depend linearly on the decision variables, the optimization problem can be converted to a quadratic programming problem and solved via off-the-shelf solvers. Once the desired robot acceleration is computed, the desired joint torques can be easily evaluated with \eqref{eq:robot_dynamics_joints}.

\begin{figure}[t]
\centering
\includegraphics[width=0.9\columnwidth]{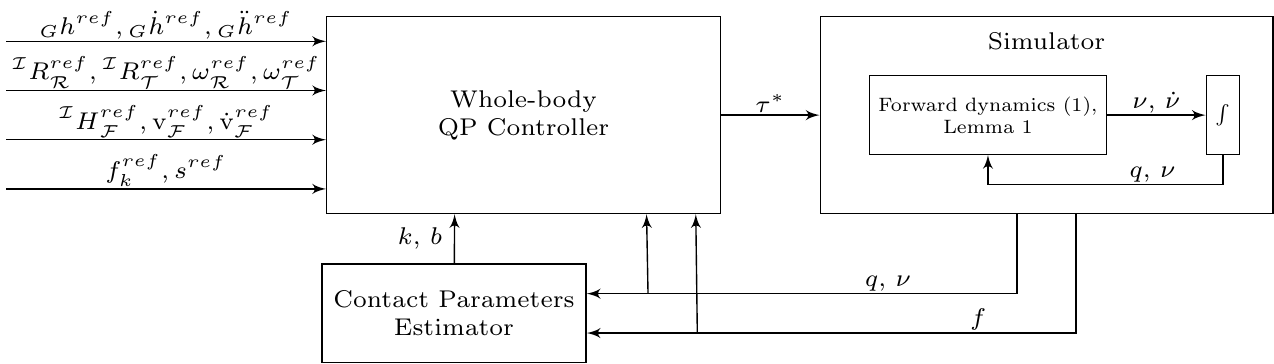}
\caption{Controller architecture.}\label{fig:block-diagram}
\end{figure}

\subsection{Contact parameters estimation}
The optimal control problem presented in Sec.~\ref{sec:tasks} is based on the perfect knowledge of the contact parameters $k$ and $b$. In a simulated environment, the value of the parameters is perfectly known. However, on the real scenario, an estimation algorithm is required to compute the parameters.
\par
The contact model described by \eqref{eq:contact_force_integral_rectangle} and \eqref{eq:contact_torque_integral_rectangle} is linear with respect the contact parameters as $f = Y(p, \prescript{\mathcal{I}}{} R _ \mathcal{F}, \dot{p}, \omega _ \mathcal{F}) \pi$,
where $Y(p, \prescript{\mathcal{I}}{} R _ \mathcal{F}, \dot{p}, \omega _ \mathcal{F}) \in \mathbb{R} ^{6 \times 2}$ is the regressor and $\pi = \begin{bmatrix} k & b \end{bmatrix} ^\top \in \mathbb{R}^2$.
We aim to estimate the contact parameters $\hat{\pi}$ such that the \emph{least-squares} criterion is minimized. Namely: 
\begin{equation}
    \label{eq:rls_definition}
    \hat{\pi}_t = \argmin\limits_{\pi} \frac{1}{2} \sum_{i=1}^t\norm{f_i - Y_i \pi}^2 _{\Gamma_i},
\end{equation}
where $\Gamma _i$ is a positive definite matrix. $f_i$ and $Y_i$ represents the forces and the regressor computed at instant $i$. 
\par
The solution of \eqref{eq:rls_definition} is given by~\cite{ljung1999system}:
\begin{equation}
    \label{eq:contact_param_est}
    \hat{\pi}_t = \hat{\pi}_{t-1} + L_t \left[f_t - Y_t \hat{\pi}_{t-1}\right],
\end{equation}
here $f_t - Y_t \hat{\pi}_{t-1}$ is the innovation. The gain $L_t \in \mathbb{R}^{2\times6}$ is:
\begin{equation}
    \label{eq:kalman}
    L_t = P_{t-1} Y_t^\top \left[ \Gamma_t + H_t P_{t-1} H^\top_t\right]^{-1},
\end{equation}
$P_{t-1}$ is the estimation error covariance at the instant $t-1$:
\begin{equation}
    \label{eq:contact_param_cov}
    P_t = \left[I_ 2- K_t H_t\right] P_{t-1} \left[I_2 - K_t H_t\right]^\top + L_t \Gamma_t  L_t^\top.
\end{equation}
At every time step, \eqref{eq:contact_param_est}-\eqref{eq:kalman}-\eqref{eq:contact_param_cov} are used to estimate the contact parameters $k$ and $b$. In the case of zero linear and angular velocity, the last three  rows of $Y$ are zero, so the damper parameter is not observable when the contact surface does not move. Walking simulations we performed, tend to show that the foot velocity is almost always different from zero when in contact with the ground. So, the contact parameters are in practice observable during robot  walking.
\section{Results \label{sec:results}}

\begin{figure*}[!t]
    \begin{myframe}{k = $\SI{2e6}{\newton \per \meter^3}$  b = $\SI{1e4}{\newton \second \per \meter^3}$}
        \begin{subfigure}[b]{0.34\textwidth}
        \centering
        \includegraphics[height=0.101\textheight]{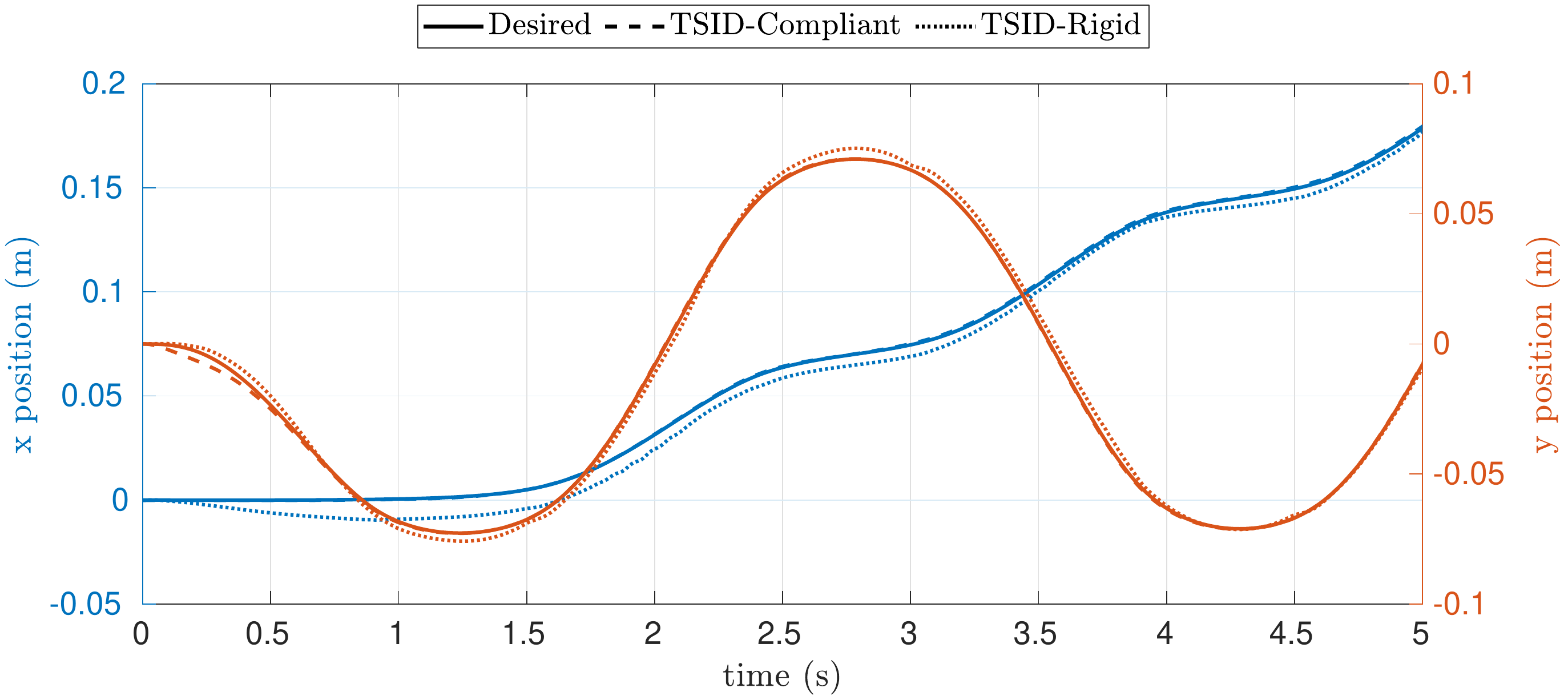}
        \caption{CoM}
        \label{fig:2e6_1e4_com}
    \end{subfigure}
    \hfill
    \begin{subfigure}[b]{0.32\textwidth}
        \centering
        \includegraphics[height=0.101\textheight]{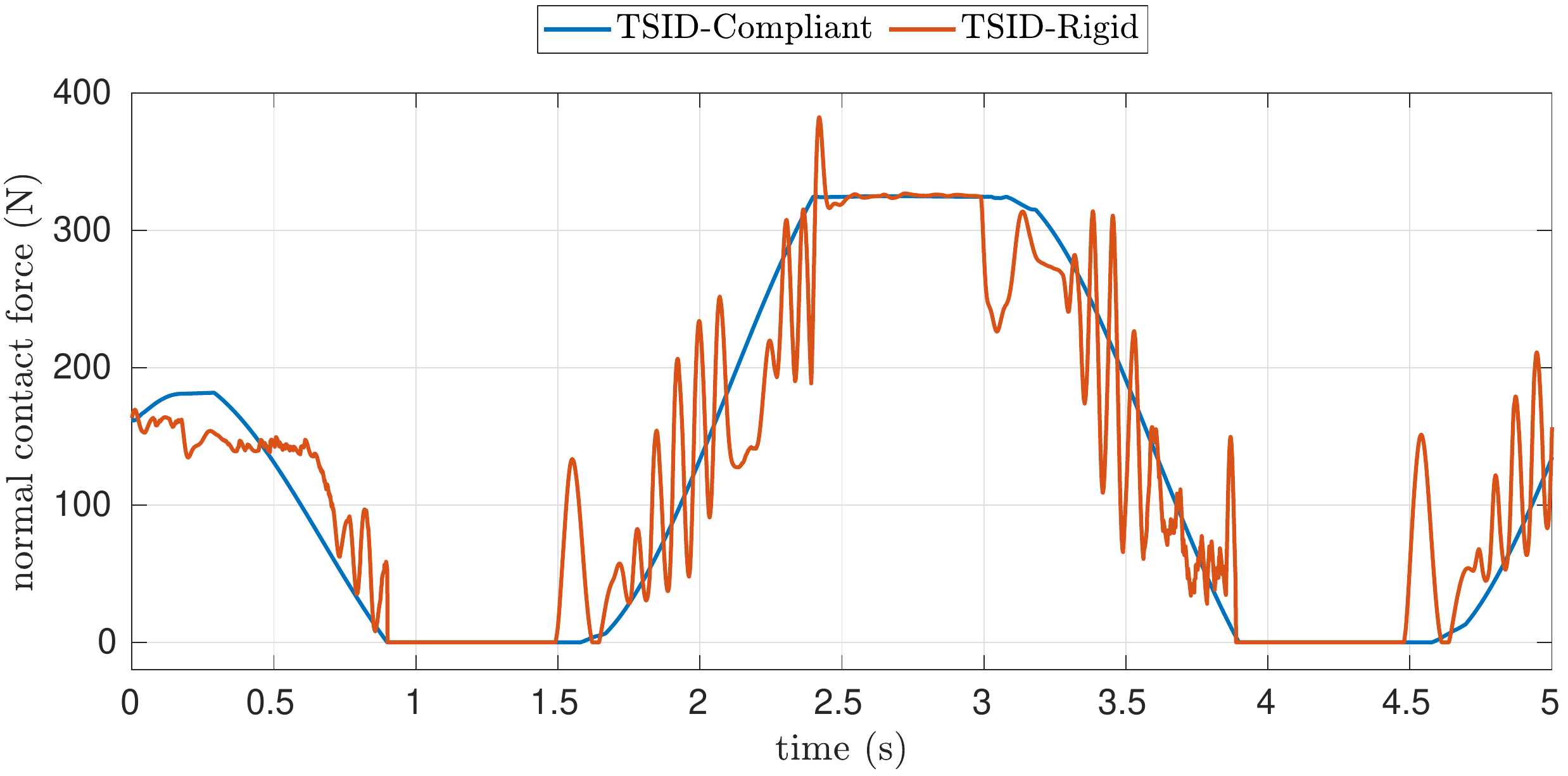}
        \caption{Normal contact force}
        \label{fig:2e6_1e4_force}
    \end{subfigure}
     \begin{subfigure}[b]{0.32\textwidth}
        \centering
        \includegraphics[height=0.101\textheight]{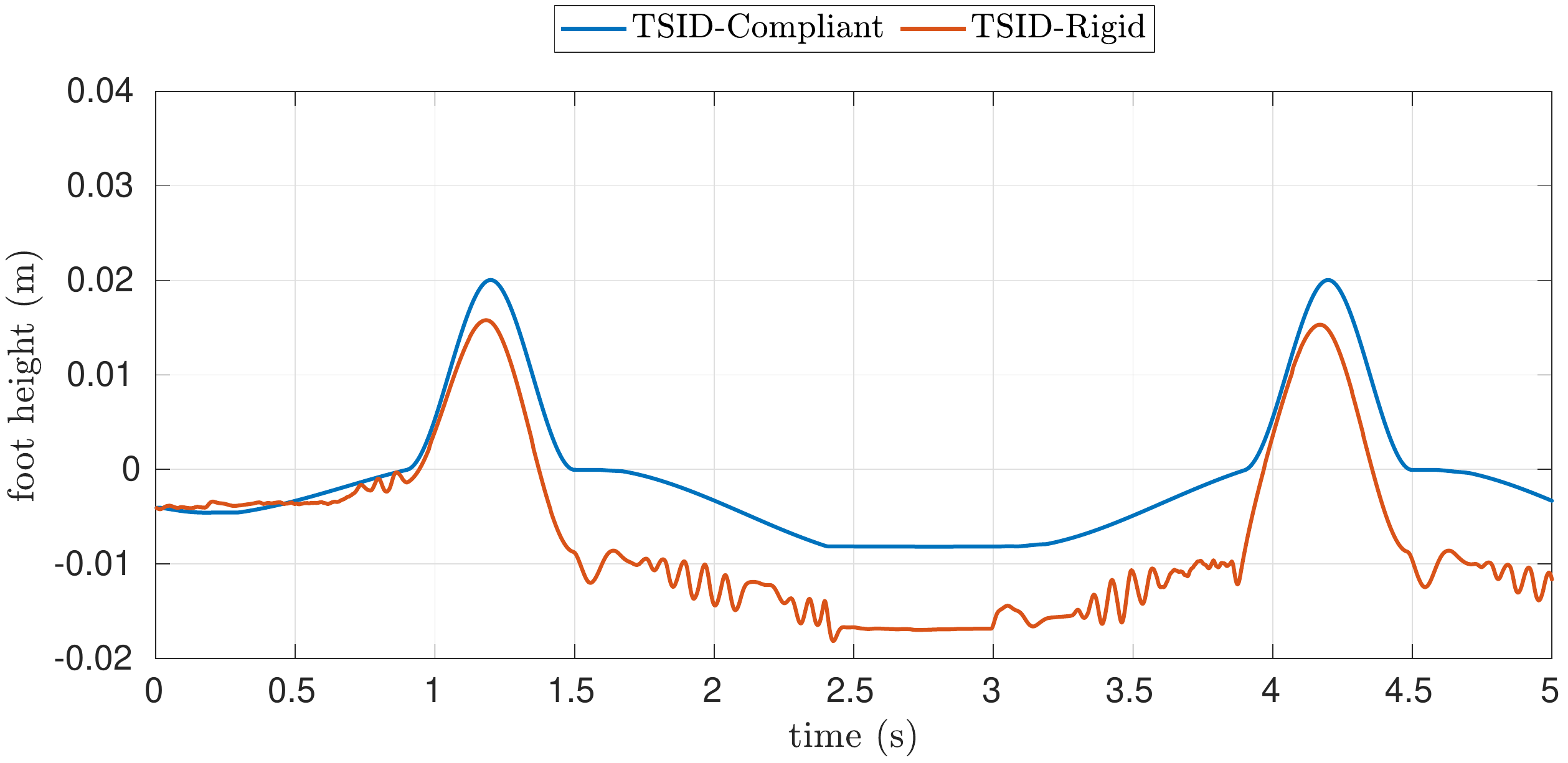}
        \caption{Foot trajectory}
        \label{fig:2e6_1e4_foot}
    \end{subfigure}
    \end{myframe}
    \caption{Comparison between TSID-Rigid and TSID-Compliant.}
\end{figure*}

\begin{figure*}[!t]
    \begin{myframe}{k = $\SI{2e6}{\newton \per \meter^3}$  b = $\SI{1e3}{\newton \second \per \meter^3}$}
        \begin{subfigure}[b]{0.34\textwidth}
        \centering
        \includegraphics[height=0.101\textheight]{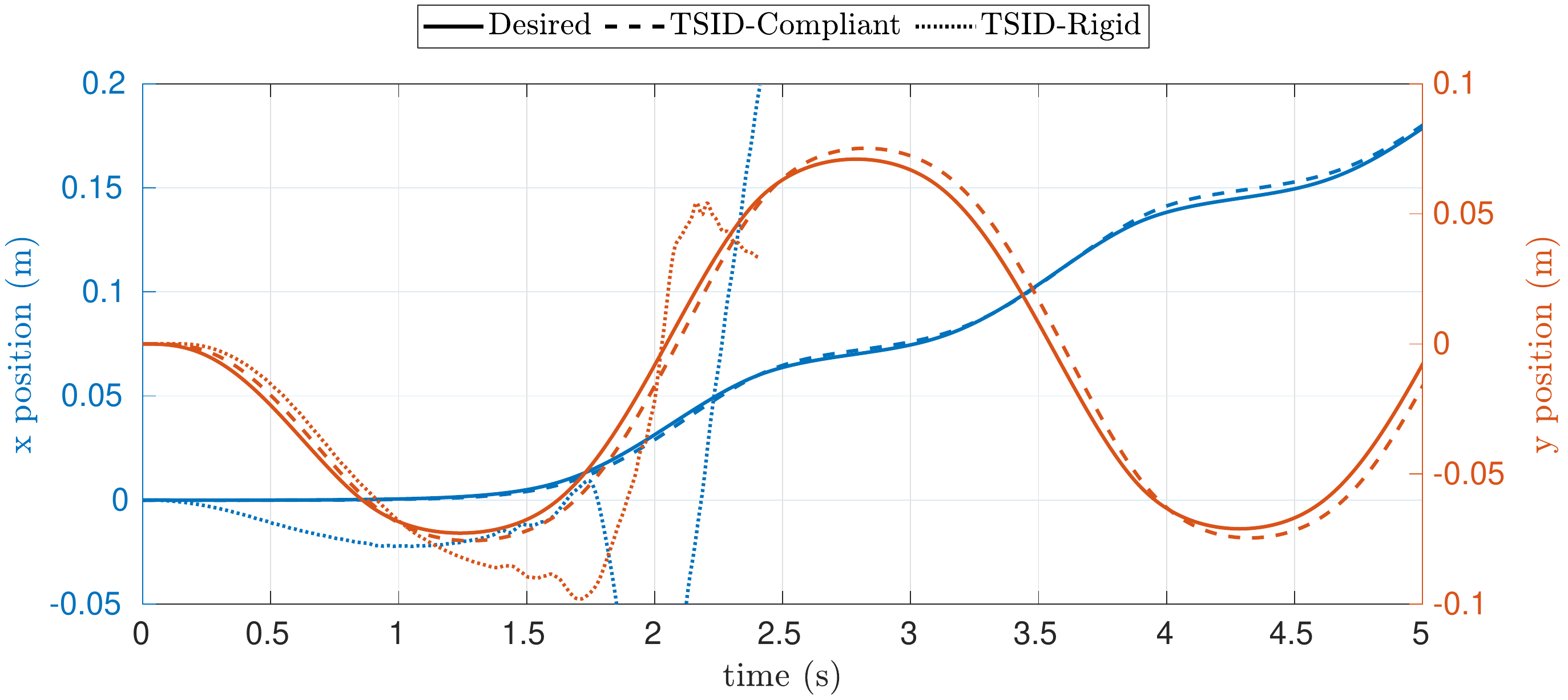}
        \caption{CoM}
        \label{fig:2e6_1e3_com}
    \end{subfigure}
    \hfill
    \begin{subfigure}[b]{0.32\textwidth}
        \centering
        \includegraphics[height=0.101\textheight]{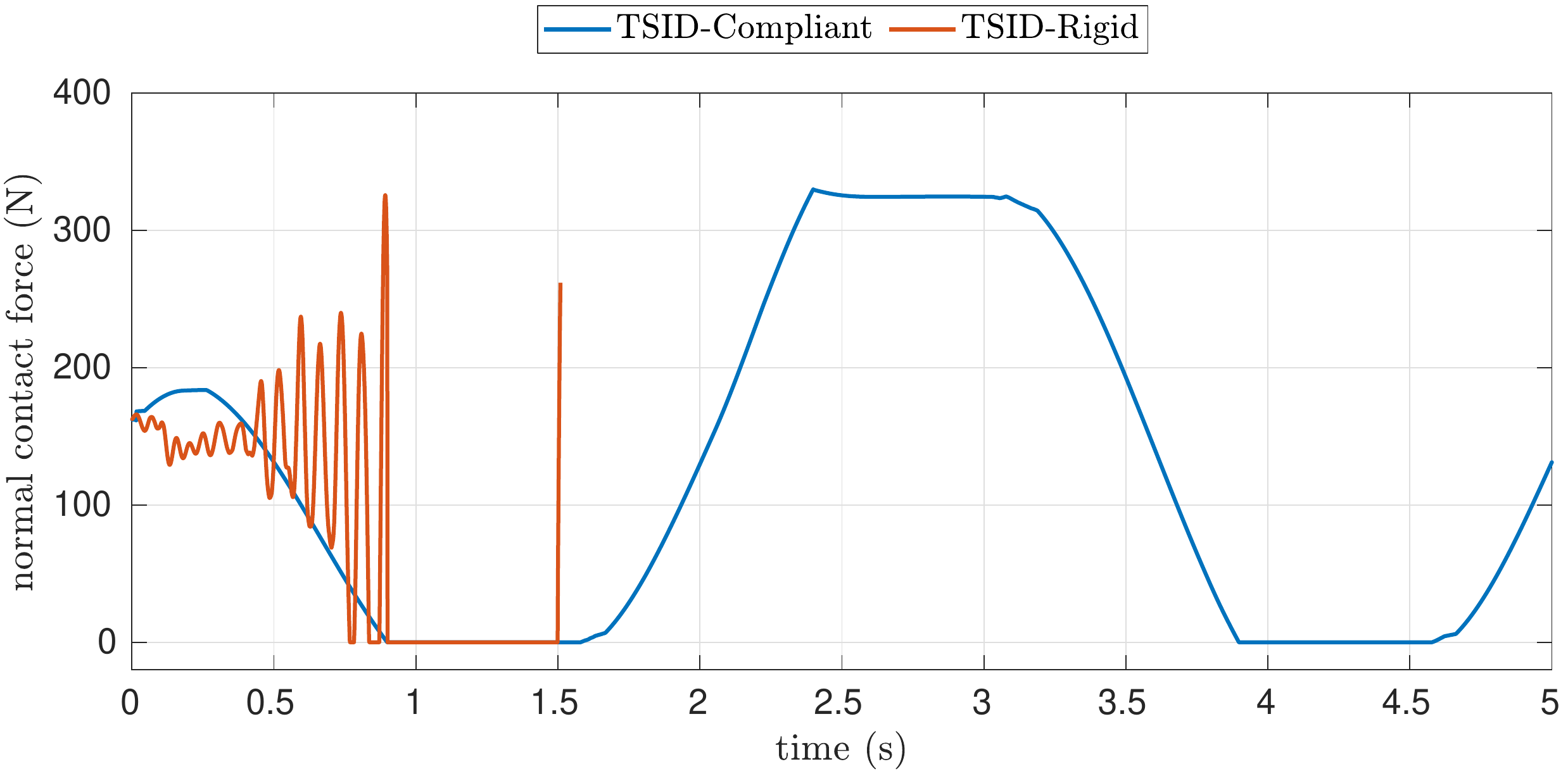}
        \caption{Normal contact force}
        \label{fig:2e6_1e3_force}
    \end{subfigure}
     \begin{subfigure}[b]{0.32\textwidth}
        \centering
        \includegraphics[height=0.101\textheight]{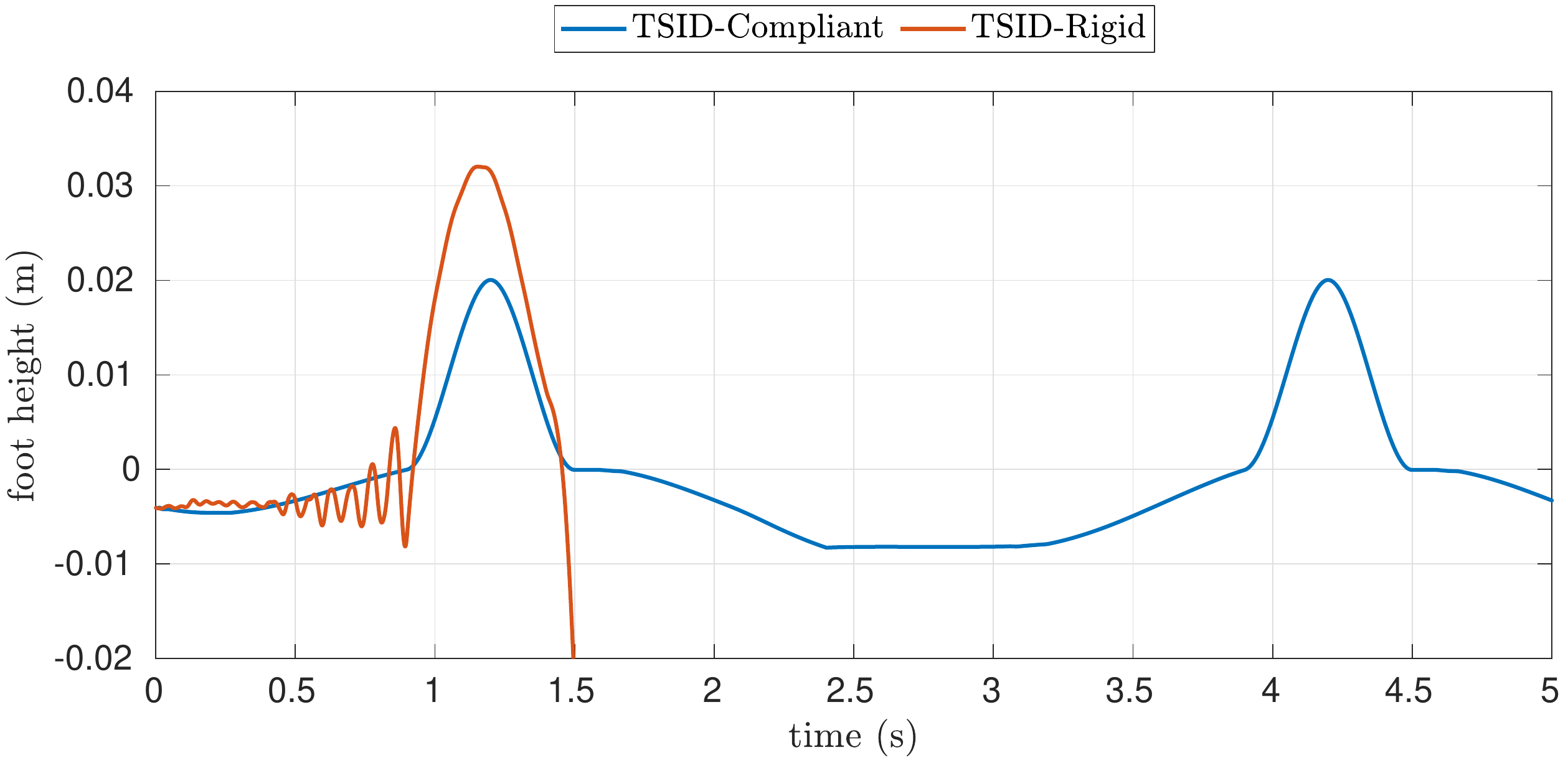}
        \caption{Foot trajectory}
        \label{fig:2e6_1e3_foot}
    \end{subfigure}
    \end{myframe}
    \caption{Comparison between TSID-Rigid and TSID-Compliant. At $t\approx \SI{1.75}{\second}$, the TSID-Rigid makes the robot fall down.}
\end{figure*}

\begin{figure*}[!t]
    \begin{myframe}{k = $\SI{1e6}{\newton \per \meter^3}$  b = $\SI{1e4}{\newton \second \per \meter^3}$}
        \begin{subfigure}[b]{0.34\textwidth}
        \centering
        \includegraphics[height=0.101\textheight]{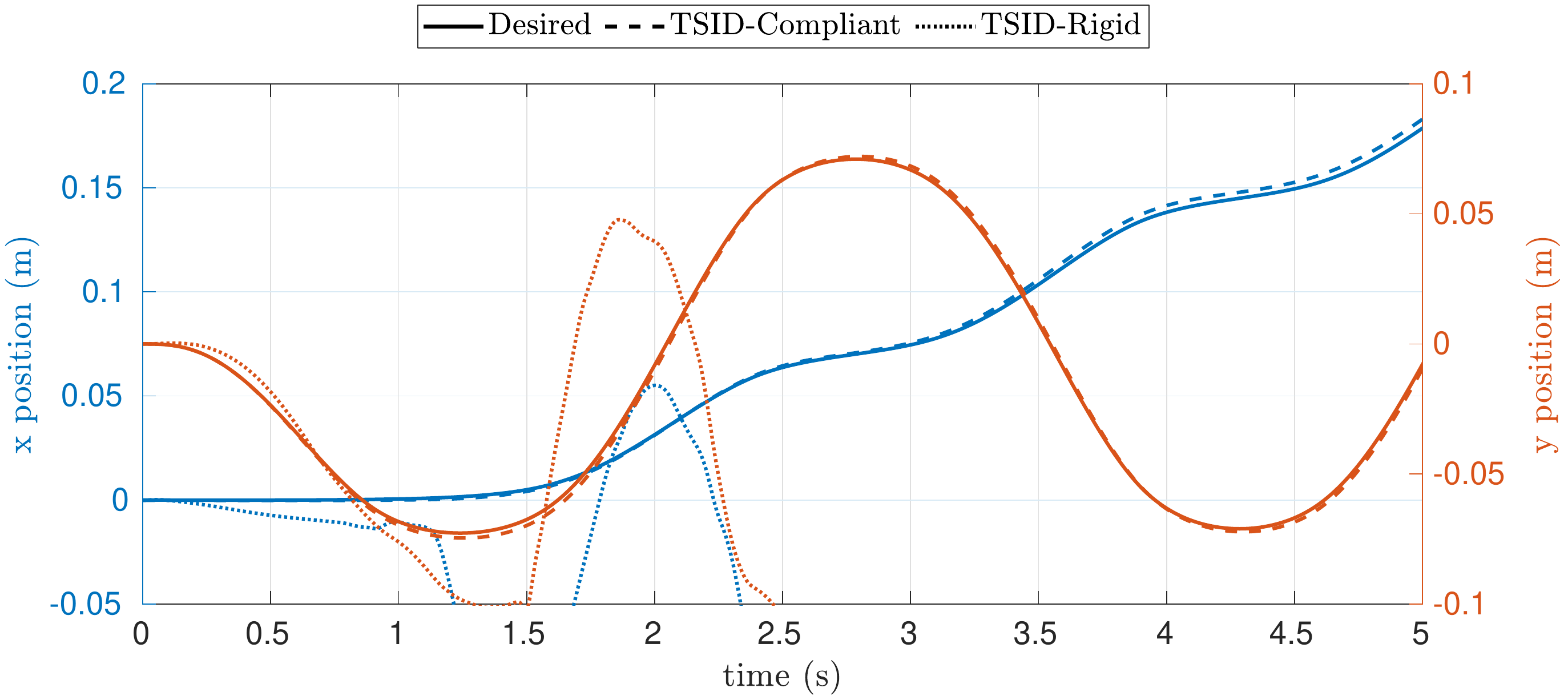}
        \caption{CoM}
        \label{fig:1e6_1e4_com}
    \end{subfigure}
    \hfill
    \begin{subfigure}[b]{0.32\textwidth}
        \centering
        \includegraphics[height=0.101\textheight]{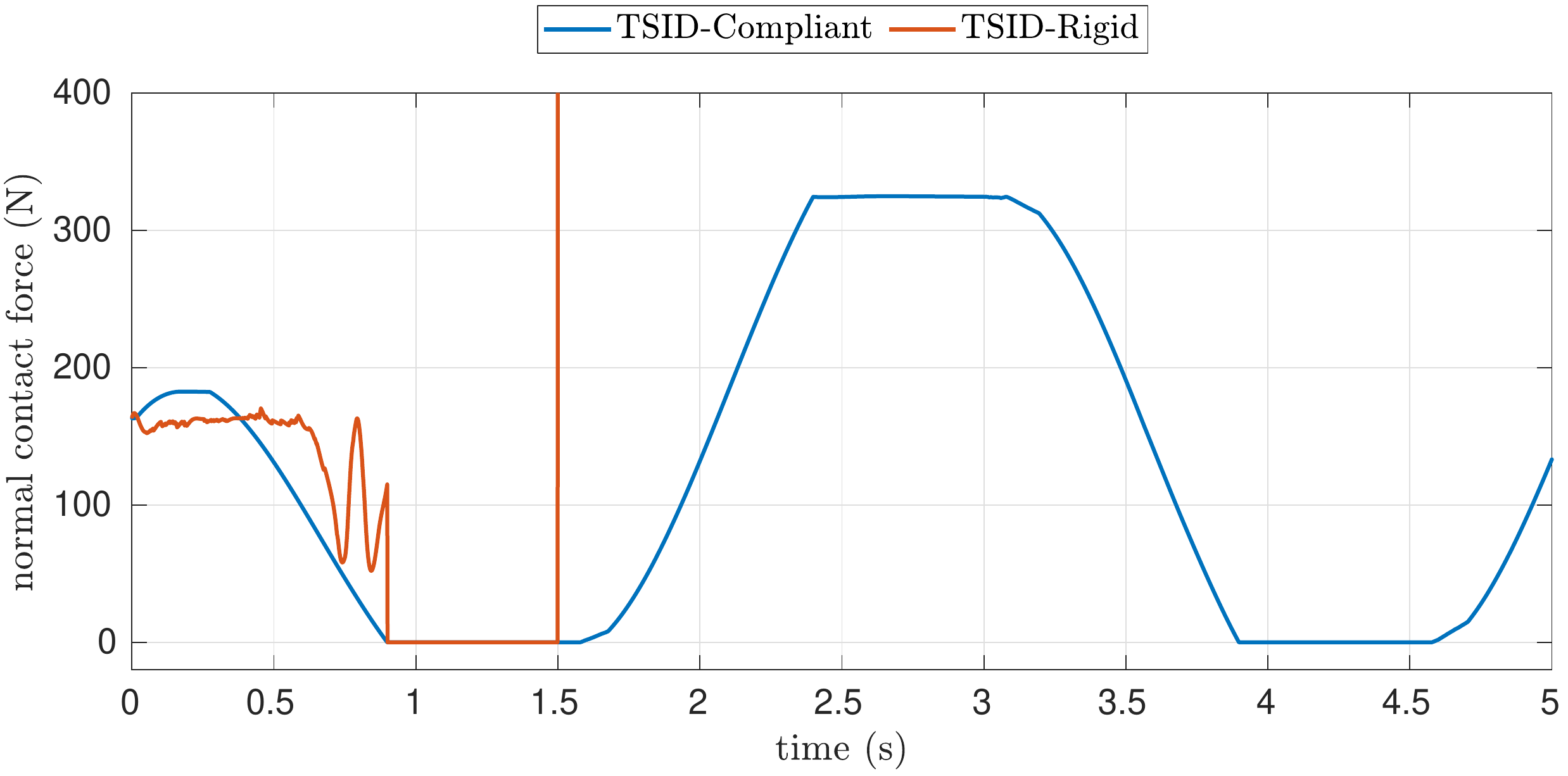}
        \caption{Normal contact force}
        \label{fig:1e6_1e4_force}
    \end{subfigure}
    \hfill
     \begin{subfigure}[b]{0.32\textwidth}
        \centering
        \includegraphics[height=0.101\textheight]{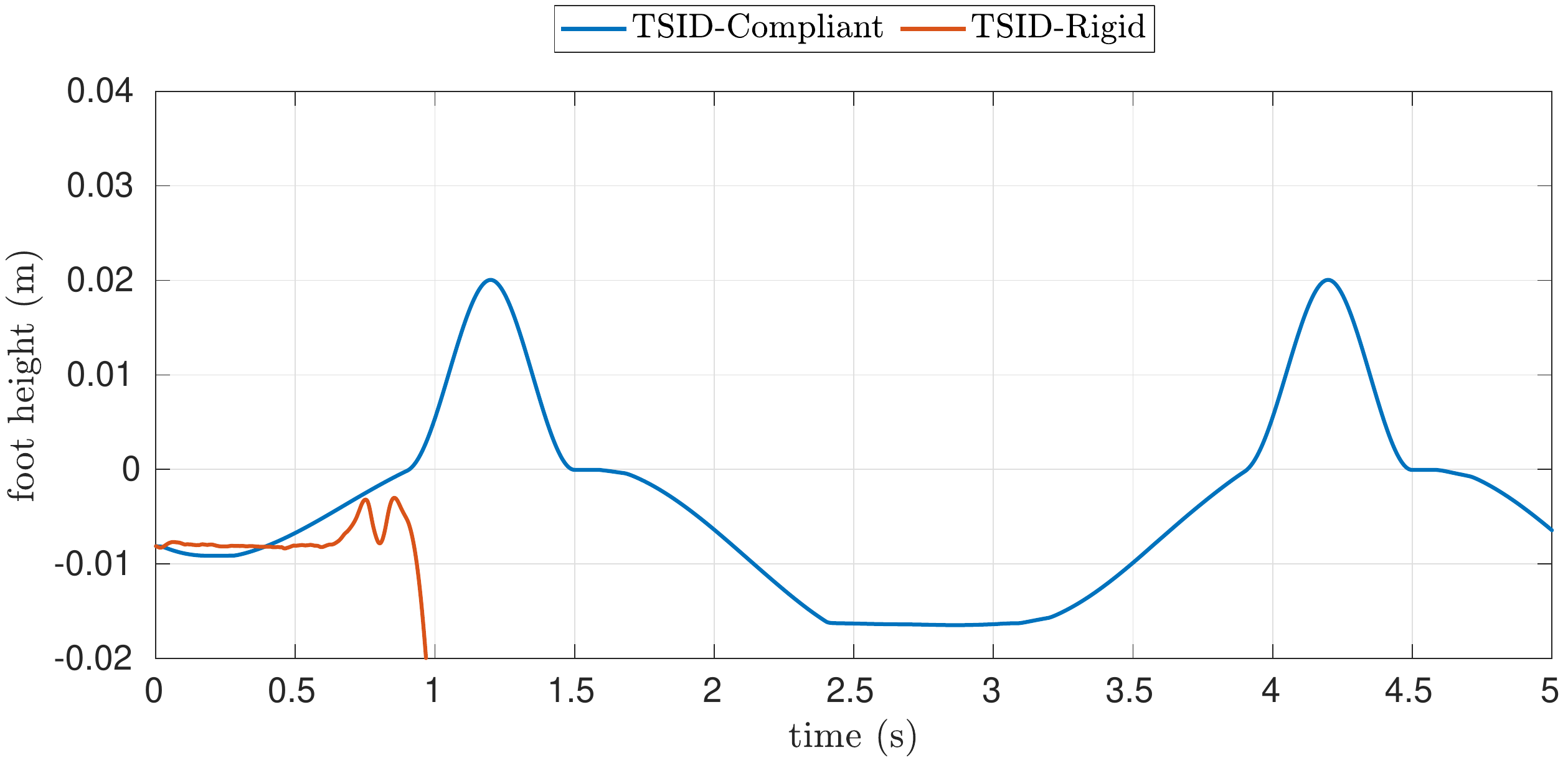}
        \caption{Foot trajectory}
        \label{fig:1e6_1e4_foot}
    \end{subfigure}
    \end{myframe}
    \caption{Comparison between TSID-Rigid and TSID-Compliant. At $t\approx \SI{0.9}{\second}$, the TSID-Rigid makes the robot fall down.}
\end{figure*}
\begin{figure*}[t]
    \begin{myframe}{$k = \SI{8e6}{\newton \per \meter^3}$ $b = \SI{1e4}{\newton \second \per \meter^3}$}
        \begin{subfigure}[b]{0.329\textwidth}
        \centering
        \includegraphics[height=0.101\textheight]{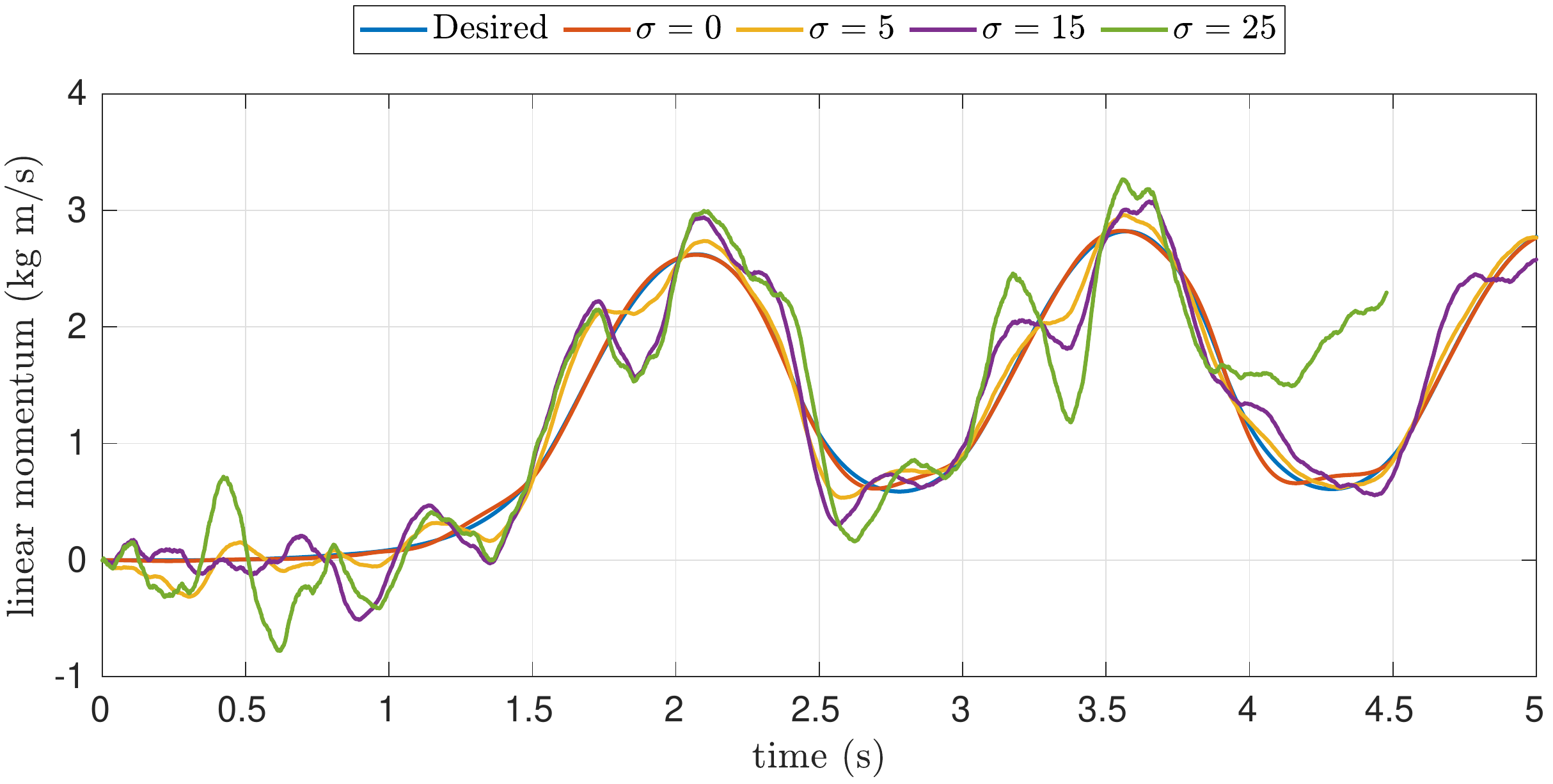}
        \caption{Linear momentum - x coordinate}
        \label{fig:noise_linear_momentum_x}
    \end{subfigure}
    \hfill
    \begin{subfigure}[b]{0.329\textwidth}
        \centering
        \includegraphics[height=0.101\textheight]{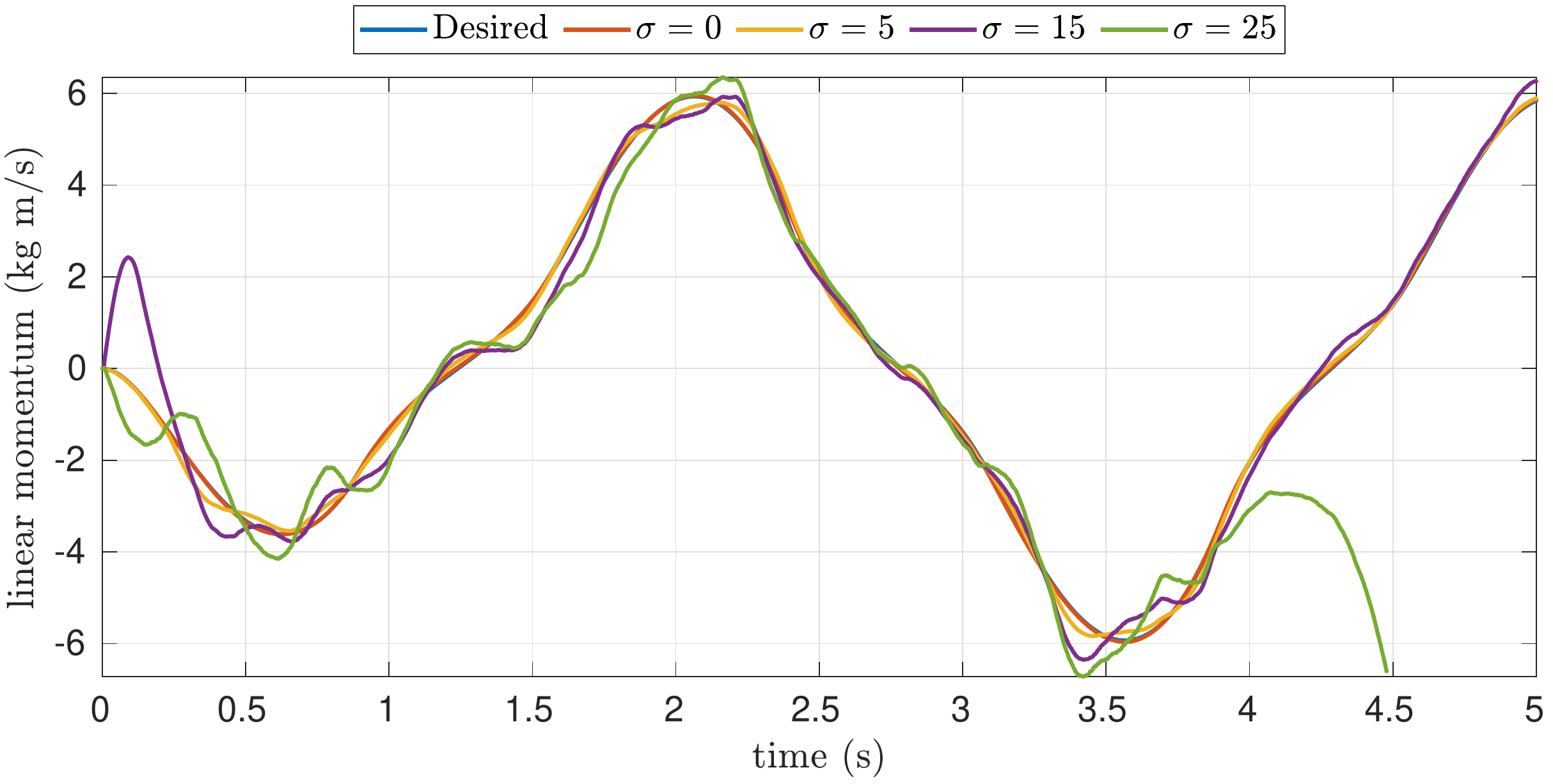}
        \caption{Linear momentum - y coordinate}
        \label{fig:noise_linear_momentum_y}
    \end{subfigure}
    \hfill
     \begin{subfigure}[b]{0.329\textwidth}
        \centering
        \includegraphics[height=0.101\textheight]{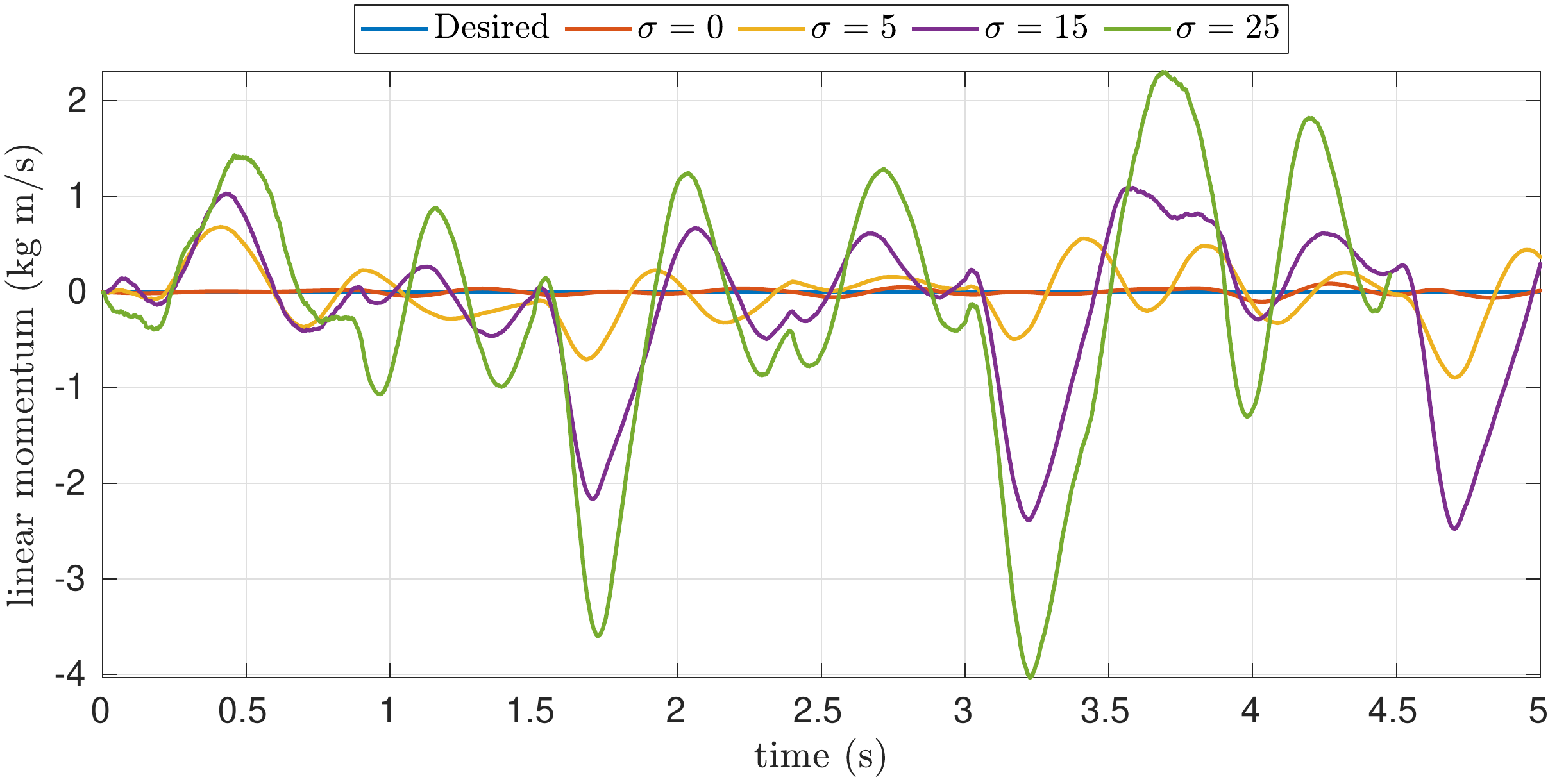}
        \caption{Linear momentum - z coordinate}
        \label{fig:noise_linear_momentum_z}
    \end{subfigure}
    \end{myframe}
    \caption{Linear momentum tracking for different values of $\sigma$. At $t\approx\SI{4}{\second}$ and $\sigma = 20$  the robot fall down. \label{fig:noise_linear_momentum}}
\end{figure*}

\begin{figure*}[!t]
    \begin{myframe}{Contact parameter estimation}
        \begin{subfigure}[b]{0.329\textwidth}
        \centering
        \includegraphics[height=0.101\textheight]{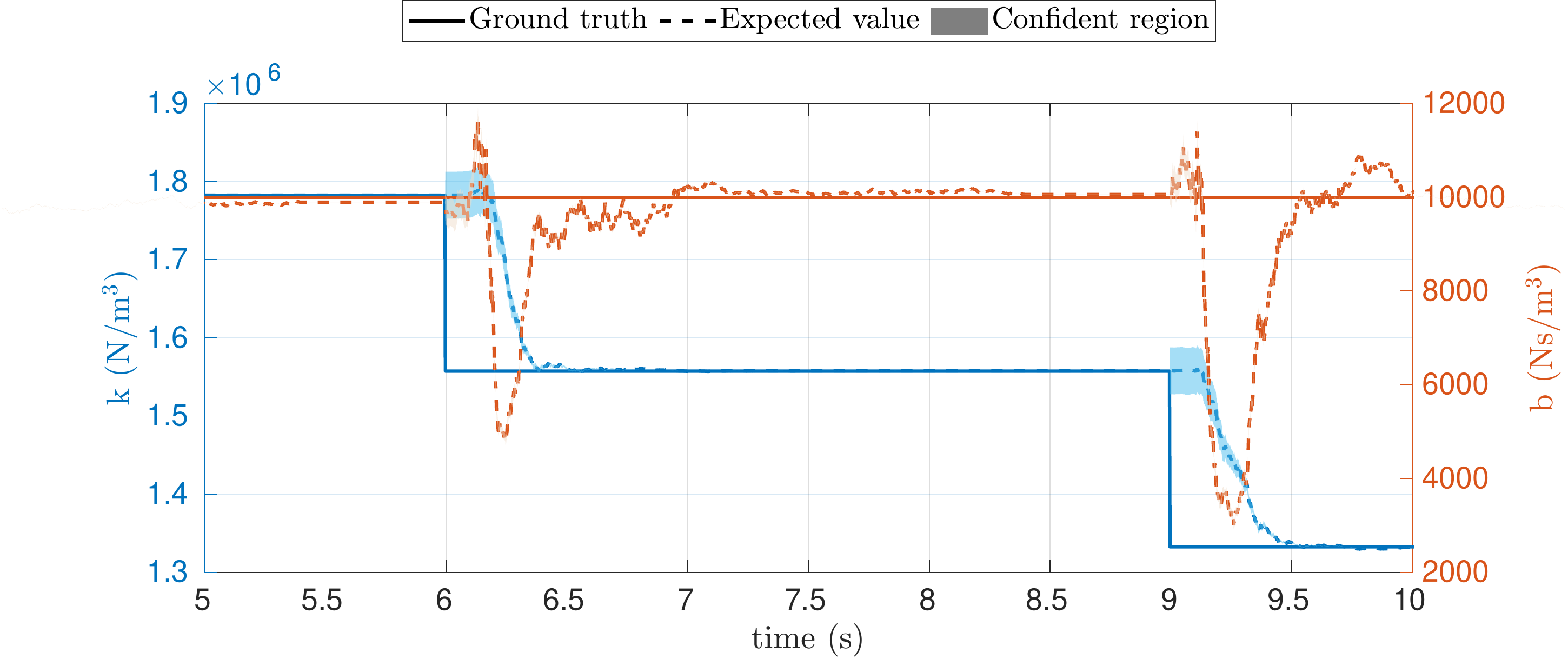}
        \caption{Fixed $b$ and space-varying $k$}
        \label{fig:compliant_spring_varing_rls}
    \end{subfigure}
    \hfill
    \begin{subfigure}[b]{0.329\textwidth}
        \centering
        \includegraphics[height=0.101\textheight]{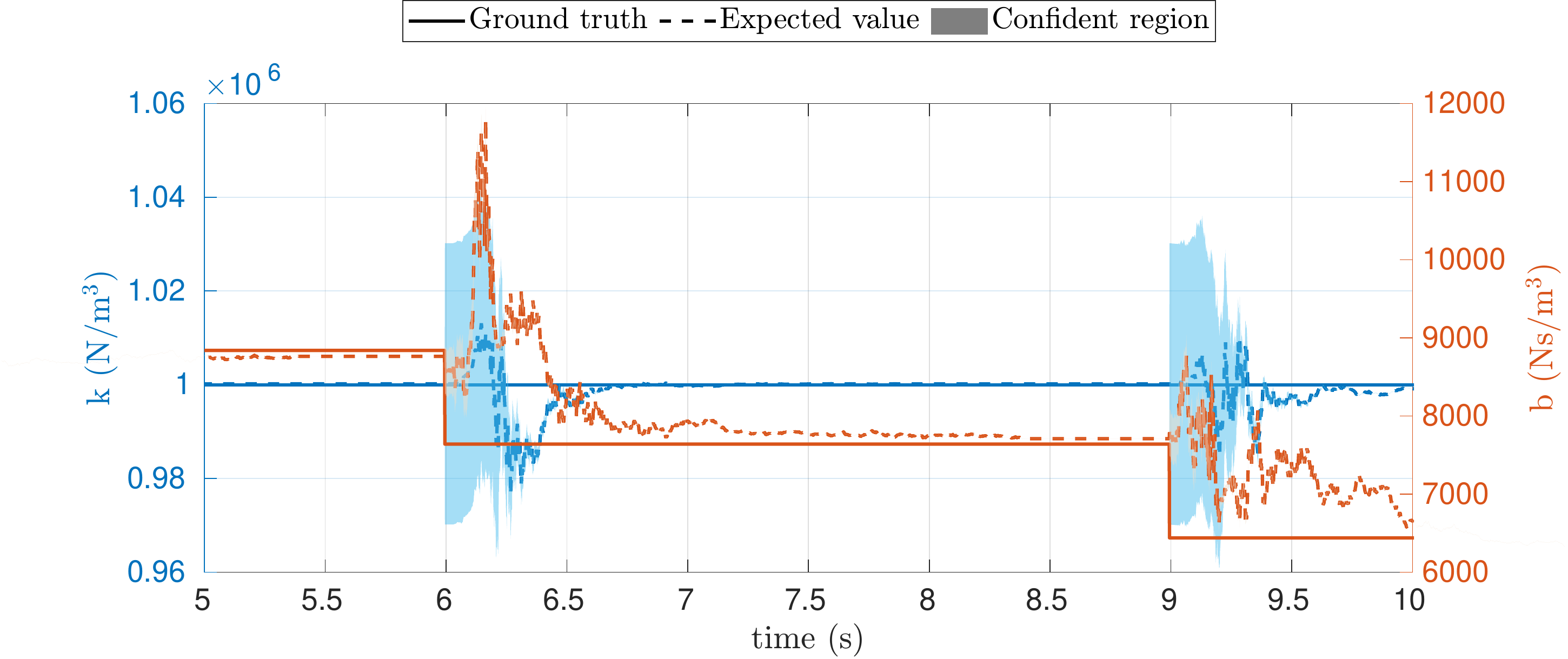}
        \caption{Fixed $k$ and space-varying $b$}
        \label{fig:compliant_damper_varing_rls}
    \end{subfigure}
     \begin{subfigure}[b]{0.329\textwidth}
        \centering
        \includegraphics[height=0.101\textheight]{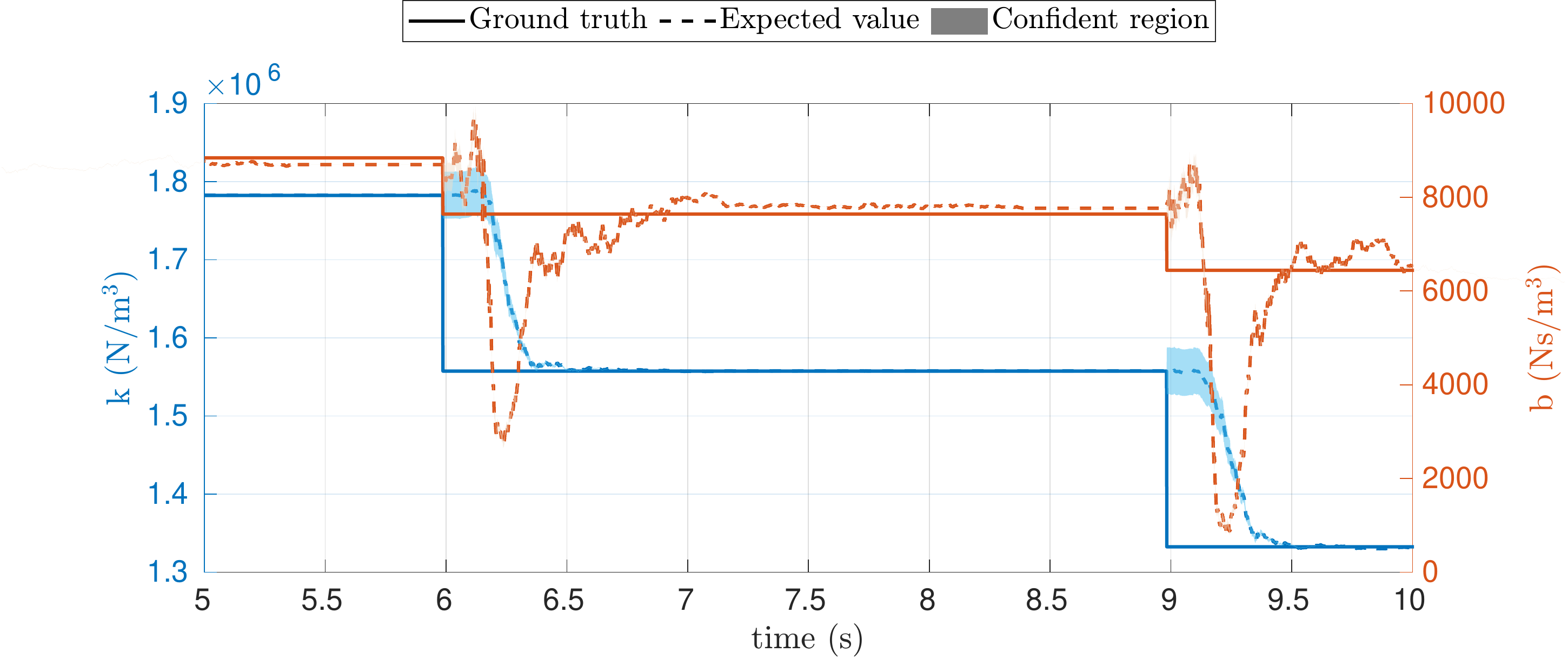}
        \caption{Space-varying $k$ and $b$}
        \label{fig:compliant_spring_damper_varing_rls}
    \end{subfigure}
    \end{myframe}
    \caption{Estimation of the contact parameters. The contact wrench is perturbed with zero-mean Gaussian noise with $\sigma = 5$.}
\end{figure*}

In this section, we test the control strategy presented in Sec~\ref{sec:controller}.
The proposed strategy 
is also compared with a state of the art task-based inverse dynamics algorithm that considers rigid contacts~\cite{doi:10.1142/S0219843619500348}. From now on, the proposed control approach is called \emph{TSID-Compliant} and the state of the art controller \emph{TSID-Rigid}.
The experiments are carried out on a simulated version of the iCub humanoid robot \cite{Metta2010}. Recall that iCub is a $\SI{104}{\centi \meter}$ tall humanoid robot. It weighs $\SI{33}{\kilo \gram}$ and it has a foot length and width of  $\SI{19}{\centi \meter}$ and $\SI{9}{\centi \meter}$, respectively. The architecture takes (in average) less than $\SI{1}{\milli \second}$ for evaluating its outputs. The OSQP~\cite{osqp} library is used to solve the optimization problems. The code is available at \url{https://github.com/dic-iit/Romualdi-2021-RAL-soft_terrain_walking}.
Simulations are obtained by integrating the robot forward dynamics (FD) obtained from~\eqref{eq:robot_dynamics}. 
The FD is evaluated with the contact model in Lemma~\ref{lemma:compliant_model} perturbed with a zero-mean Gaussian noise.

To validate the performances of the proposed architecture, we present three main experiments.
First, we compare the performances of the TSID-Compliant and the TSID-Rigid controller in case of different contact parameters. Secondly, we analyze the robustness of the TSID-Compliant in case of non-parametric uncertainty in the contact model. Finally, we show the contact parameter estimation performances in case of anisotropic environment. In all the scenarios, the robot walks straight and the maximum velocity is $\SI{0.17}{\meter \per \second}$.

\subsection{Comparison between TSID-Compliant and TSID-Rigid}
Tab.~\ref{tab:architecture_outcome} summarizes the outcome of the control strategies for different contact parameters. The labels \emph{success} and \emph{failure} mean that the associated controller is either able or not to ensure the robot balancing while walking.
\par
To compare the two controllers, we decided to perform three main experiments. In the first experiment, we choose a set of contact parameters such that both  whole-body controllers guarantee the balance while walking. In the second one, we keep the damper coefficient $b$ constant, and we decrease the spring coefficient $k$. Finally, in the third experiment we keep $k$ constant, and we decreased $b$.  Namely: 
\begin{itemize}
    \item[\textbf{-Experiment 1}] $k=\SI{2e6}{\newton \per \meter^3}$, $b = \SI{1e4}{\newton \second \per \meter^3}$;
    \item[\textbf{-Experiment 2}] $k=\SI{1e6}{\newton \per \meter^3}$, $b=\SI{1e4}{\newton \second \per \meter^3}$;
    \item[\textbf{-Experiment 3}] $k = \SI{2e6}{\newton \per \meter^3}$, $b=\SI{1e3}{\newton \second \per \meter^3}$.
\end{itemize}

\begin{table}[t]
    \centering
    \caption{Controller  implementation outcomes.
    }
    \begin{tabular}{|c|c|c|c|}
    \hline \rowcolor[gray]{.9}
         \begin{tabular}{@{}c@{}}TSID Type\end{tabular} &
         \begin{tabular}{@{}c@{}}k ($\SI{}{\newton \per \meter^3}$)\end{tabular} &
         \begin{tabular}{@{}c@{}}b ($\SI{}{\newton \second \per \meter^3}$)\end{tabular} &
         \begin{tabular}{@{}c@{}}Outcome\end{tabular}\\
        \hline
        Compliant  & $\SI{8e5}{}$  &  $\SI{1e4}{}$ & Success \\
        \hline
        Compliant  & $\SI{1e6}{}$  &  $\SI{1e4}{}$ & Success \\
        \hline
        Compliant  & $\SI{2e6}{}$  &  $\SI{1e4}{}$ & Success \\
        \hline
        Compliant  & $\SI{8e5}{}$  &  $\SI{1e3}{}$ & Success \\
        \hline
        Compliant  & $\SI{2e6}{}$  &  $\SI{1e3}{}$ & Success \\
        \hline
        Compliant  & $\SI{1e6}{}$  &  $\SI{1e3}{}$ & Success \\
        \hline
        Rigid  & $\SI{8e5}{}$  &  $\SI{1e4}{}$ & Failure \\
        \hline
        Rigid  & $\SI{1e6}{}$  &  $\SI{1e4}{}$ & Failure \\
        \hline 
        Rigid  & $\SI{2e6}{}$  &  $\SI{1e4}{}$ & Success \\
        \hline 
        Rigid  & $\SI{8e5}{}$  &  $\SI{1e3}{}$ & Failure \\
        \hline 
        Rigid  & $\SI{1e6}{}$  &  $\SI{1e3}{}$ & Failure \\
        \hline
        Rigid  & $\SI{2e6}{}$  &  $\SI{1e3}{}$ & Failure \\
        \hline

    \end{tabular}
    \label{tab:architecture_outcome}
\end{table}

\subsubsection{Experiment 1} 
Figure~\ref{fig:2e6_1e4_com}  depicts the CoM tracking performance obtained with the TSID-Compliant and the TSID-Rigid. Both controllers seem to show good tracking performances and the CoM error is kept below $\SI{1}{\centi \meter}$ in both cases. Note that the TSID-Rigid controller induces faster variations on the measured contact wrenches -- Figure~\ref{fig:2e6_1e4_force}. This contributes to overall higher vibrations of the robot. One reason for this behaviour is that the TSID-Rigid assumes full control on the desired contact wrenches. This assumption is in general valid in case of stiff contacts, but it does not hold if the environment is compliant. Figure~\ref{fig:2e6_1e4_foot} presents the left foot trajectory when the whole-body controller is either TSID-Compliant or TSID-Rigid. The TSID-Compliant ensures a smoother foot motion when it is in contact with the environment ($\SI{1.5}{\second}<t<\SI{3.5}{\second}$).

\subsubsection{Experiment 2} 
There is no significant difference between the CoM tracking obtained with the two implementations of the whole-body controller -- Figure~\ref{fig:2e6_1e3_com} for $t < \SI{1}{\second}$. However, the lower the damper parameter $b$, the higher is the acceleration required to change the contact wrench. This goes in contrast with the assumption of zero foot accelerations of the TSID-Rigid controller. Consequentially, the TSID-Rigid generates fast vibrations of the measured contact wrench and on the foot position -- Figures~\ref{fig:2e6_1e3_force} and \ref{fig:2e6_1e3_foot} respectively. Clearly, these bad performances in turn induce a bad tracking of the CoM shown in Fig.~\ref{fig:2e6_1e3_com} at $t\approx \SI{1.5}{\second}$, and consequently the robot fall down. A possible solution to mitigate this problem is to increase the CoM gain in the TSID-Rigid controller. This unfortunately gives rise to higher robot oscillations, which in turn degrade the CoM tracking, and still the robot falls or, even worst, the controller may not be able to find a feasible solution.
\subsubsection{Experiment 3} 
The CoM tracking problem presented in the Experiment 2 worsens at lower values of the contact parameter $k$.
Figure~\ref{fig:1e6_1e4_com} shows the CoM tracking performances of the two controllers. The TSID-Compliant is still able to guarantee good performances. On the other hand, the TSID-Rigid generates faster variations on the measured contact wrenches and foot positions -- Figures~\ref{fig:2e6_1e4_force} and ~\ref{fig:2e6_1e4_foot}, respectively. Consequentially, the controller, in order to keep the balance, requires high variations of the joint accelerations, and at $t\approx\SI{0.9}{\second}$ fails while searching feasible joint torques. Although tuning the TSID-Rigid controller may mitigate this problem, we were only able to postpone the failure. 

\subsection{Robustness of the TSID-Compliant}
In this section, we present the robustness capabilities of the TSID-Compliant controller in case of non-parametric uncertainty in the contact model.

We model the non-parametric uncertainty by using an \emph{additive white Gaussian noise}. So, the contact wrench acting on the robot becomes $f_k^m = f_k + \epsilon$, where $\epsilon$ is sampled from a Gaussian distribution with zero mean and standard deviation $\sigma$. $f$ is the contact wrench computed with the contact model~\eqref{eq:contact_model}.
Fig.~\ref{fig:noise_linear_momentum} shows the linear momentum tracking performance obtained with different values of $\sigma$. The experiments are performed in case of $k = \SI{8e6}{\newton \per \meter^3}$ and $b = \SI{1e4}{\newton \second \per \meter^3}$, however similar considerations hold for the other sets of parameters -- see Tab.~\ref{tab:architecture_outcome}. 
The higher $\sigma$, the higher is the tracking error. The controller is able to guarantee good tracking performances for every $\sigma < 20$. Indeed at $\sigma = 20$ the controller is no more able to guarantee acceptable performances and at $t\approx\SI{4}{\second}$ the robot falls down.

\subsection{Anisotropic environment}
In this section, we show the contact parameter estimator performances in the case of an anisotropic environment. 
We model the contact parameters as a piece-wise function of the forward walking direction assuming that all points of the contact surface have the same contact parameters. Also, to further test the estimator robustness, we add a zero-mean Gaussian noise (with standard deviation $\sigma = 5$) to the contact wrench.
To cope with contact parameter discontinuities, we reset the state co-variance in the recursive least-square algorithm \cite{4054664}. Each time a foot impacts the ground, the co-variance is reset to the initial value. 
Fig.~\ref{fig:compliant_spring_varing_rls} shows the  estimator performance in the case of $b=\SI{1E4}{\newton\second\per\meter^3}$ and a space-dependent $k$. The discontinuity on the ground truth ($t\approx\SI{6}{\second}$ and $t\approx\SI{9}{\second}$) occurs when the swing foot makes contact with environment. The observed parameters converge to the ground truth in less than a second.
The accompanying video shows that during the transient phases (e.g. $t\approx\SI{6.5}{\second}$), the controller achieves good overall performances.
Similar considerations hold in the case $k=\SI{1e6}{\newton\per\meter^3}$ and varying $b$ -- Fig.~\ref{fig:compliant_damper_varing_rls} and space varying $k$ and $b$ -- Fig.~\ref{fig:compliant_spring_damper_varing_rls}.
\section{Conclusions \label{sec:conclusions}}
This paper contributes towards the development of a contact model to represent the interaction between the robot and the environment. Unlike state of the art models, we consider the environment as a continuum of spring-damper systems. This allows us to compute the equivalent contact wrench by integrating the pressure distribution over the contact surface. As a result, rotational springs and dampers are not required to model the interaction between the robot and the environment. We also develop a whole-body controller that stabilizes the robot walking in a compliant environment. Finally, an estimation algorithm is also introduced to compute the contact parameters in real-time. The proposed controller is then compared with a state of the art whole-body controller. We analyze the robustness properties of the architecture with respect to non-parametric uncertainty in the contact model. We finally study the performances of the contact parameter estimation in case of anisotropic environment. As a future work, we plan to embed the contact parameter estimation inside the controller, thus increasing its robustness properties against parameter uncertainties, without the need of an additional estimation algorithm. We also plan to draw a detailed comparison with other state-of-the-art controllers that assume compliant environment.
In addition, we plan to validate the architecture on the real robot. Here the performances of the low-level torque control along with the base estimation capabilities become pivotal to obtain successfully results. \looseness=-1 
\appendix
\label{sec:appendix_compliant_model}

Let assume that each point in the contact surface is subject to an infinitesimal force $\rho \in \mathbb{R}^3$  \eqref{eq:contact_model_general}. 
Given an inertial frame $\mathcal{I}$ and a frame $\mathcal{F}$ rigidly attached to the body, each point $x$ on the contact surface is described by
\begin{equation}
x = p + \prescript{\mathcal{I}}{} R _ \mathcal{F} \begin{bmatrix} u & v & 0 \end{bmatrix}^\top,
\label{eq:contact_point_position}
\end{equation}
where $p \in \mathbb{R}^3$ is the origin of the frame $\mathcal{F}$ w.r.t. the inertial frame $\mathcal{I}$, $\prescript{\mathcal{I}}{}R _\mathcal{F}$ is the rotation from $\mathcal{I}$ to $\mathcal{F}$. $u$ and $v$ are the coordinates, in the foot frame $\mathcal{F}$, of the point laying into the contact surface.
The point position time derivative is: 
\begin{equation}
\dot{x} = \dot{p} + S( {}^{\mathcal{I}} \omega _ {\mathcal{F}}) {}^\mathcal{I} R _ \mathcal{F} \begin{bmatrix} u & v & 0 \end{bmatrix} ^\top,
\label{eq:contact_point_velocity}
\end{equation}
Using the hypothesis of rigid-body, $\bar{x}$ can be computed as:
\begin{equation}
\bar{x} = \bar{p} + {}^\mathcal{I} \bar{R} _ \mathcal{F} \begin{bmatrix} u & v & 0 \end{bmatrix} ^\top.
\label{eq:null_force_point}
\end{equation}
Here $\bar{p}$ and ${}^\mathcal{I} \bar{R} _ \mathcal{F}$ are the position and rotation of the body frame associated to a null force.
\par
Combining \eqref{eq:contact_model_general} with \eqref{eq:contact_point_position}, \eqref{eq:contact_point_velocity} and \eqref{eq:null_force_point}, the force acting on a point lying on the contact surface becomes:
\begin{IEEEeqnarray}{LL}
\IEEEyesnumber \phantomsection
\label{eq:contact_model}
\rho &= k \left\{ \bar{p} - p + (\prescript{\mathcal{I}}{}{\bar{R}}_\mathcal{F} - \prescript{\mathcal{I}}{}R _\mathcal{F}) \begin{bmatrix} u & v & 0 \end{bmatrix}^\top \right\} \IEEEyessubnumber \label{eq:contact_model_spring}\\
&- b \left\{ \dot{p} + S({}^{\mathcal{I}} \omega _ {\mathcal{F}}) \prescript{\mathcal{I}}{}R _\mathcal{F} \begin{bmatrix} u & v & 0 \end{bmatrix}^\top \right\}. \IEEEyessubnumber \label{eq:contact_model_damper}
\end{IEEEeqnarray}
where \eqref{eq:contact_model_spring} is the force generated by the spring and \eqref{eq:contact_model_damper} the one produced by the damper. 
To facilitate the process of finding the solutions to the integrals~\eqref{eq:contact_wrench_generic}, let us remind that given
a double integral of a function $g(x, y)$, a variable change of the form \eqref{eq:contact_point_position}, \eqref{eq:contact_point_velocity} and \eqref{eq:null_force_point} yields
\begin{equation}
    \int\!\!\!\! \int g(x, y) \diff x \diff y = \int\!\!\!\! \int g(x(u,v), y(u,v)) |\det(J)| \diff u \diff v
    \label{eq:integral_rule}
\end{equation}
where $J$ is the Jacobian of the variable transformation. 
\par
\eqref{eq:integral_rule} and \eqref{eq:contact_model} can be used to evaluate the total force applied from the environment to a generic contact surface. If $\Omega$ is represented by a rectangle with a length $l$ and a width $w$ $J = |e _ 3^\top \prescript{\mathcal{I}}{}R _\mathcal{F} e _ 3|$ and $f _ l$ is given by~\eqref{eq:contact_force_integral_rectangle}.

The torque about the origin of $\mathcal{F}$ produced by $\rho$ is:
\begin{equation}
\label{eq:contact_torque}
\mu(u,v) = S\left(\prescript{\mathcal{I}}{}R _\mathcal{F} \begin{bmatrix} u &
 v &
 0 \end{bmatrix} ^\top
 \right) \rho(u,v)
\end{equation}
By applying~\eqref{eq:integral_rule}, the integral of~\eqref{eq:contact_torque} over a rectangular contact surface leads to~\eqref{eq:contact_torque_integral_rectangle}.

\bibliography{locomotion}

\begin{thebibliography}{10}
\providecommand{\url}[1]{#1}
\csname url@rmstyle\endcsname
\providecommand{\newblock}{\relax}
\providecommand{\bibinfo}[2]{#2}
\providecommand\BIBentrySTDinterwordspacing{\spaceskip=0pt\relax}
\providecommand\BIBentryALTinterwordstretchfactor{4}
\providecommand\BIBentryALTinterwordspacing{\spaceskip=\fontdimen2\font plus
\BIBentryALTinterwordstretchfactor\fontdimen3\font minus
  \fontdimen4\font\relax}
\providecommand\BIBforeignlanguage[2]{{%
\expandafter\ifx\csname l@#1\endcsname\relax
\typeout{** WARNING: IEEEtran.bst: No hyphenation pattern has been}%
\typeout{** loaded for the language `#1'. Using the pattern for}%
\typeout{** the default language instead.}%
\else
\language=\csname l@#1\endcsname
\fi
#2}}

\bibitem{Koolen2016}
T.~Koolen, S.~Bertrand, G.~Thomas, T.~{De Boer}, T.~Wu, J.~Smith,
  J.~Englsberger, and J.~Pratt, ``{Design of a Momentum-Based Control Framework
  and Application to the Humanoid Robot Atlas},'' \emph{Int. J. Humanoid
  Robot.}, 2016.

\bibitem{feng2015optimization}
S.~Feng, E.~Whitman, X.~Xinjilefu, and C.~G. Atkeson, ``{Optimization-based
  Full Body Control for the DARPA Robotics Challenge},'' \emph{J. F. Robot.},
  vol.~32, no.~2, pp. 293--312, 2015.

\bibitem{Kajita2001}
S.~Kajita, F.~Kanehiro, K.~Kaneko, K.~Yokoi, and H.~Hirukawa, ``{The 3D linear
  inverted pendulum model: a simple modeling for biped walking pattern
  generation},'' \emph{Proc. 2001 IEEE/RSJ Int. Conf. Intell. Robot. Syst.},
  no. October 2016, pp. 239--246, 2001.

\bibitem{Englsberger2011}
J.~Englsberger, C.~Ott, M.~A. Roa, A.~Albu-Sch{\"{a}}ffer, and G.~Hirzinger,
  ``{Bipedal walking control based on capture point dynamics},'' in \emph{IEEE
  Int. Conf. Intell. Robot. Syst.}, 2011, pp. 4420--4427.

\bibitem{Englsberger2013}
J.~Englsberger, C.~Ott, and A.~Albu-Schaffer, ``{Three-dimensional bipedal
  walking control using Divergent Component of Motion},'' in \emph{IEEE Int.
  Conf. Intell. Robot. Syst.}, 2013, pp. 2600--2607.

\bibitem{nava16}
G.~Nava, F.~Romano, F.~Nori, and D.~Pucci, ``{Stability Analysis and Design of
  Momentum-based Controllers for Humanoid Robots},'' \emph{Intell. Robot. Syst.
  2016. IEEE Int. Conf.}, 2016.

\bibitem{doi:10.1142/S0219843619500348}
G.~Romualdi, S.~Dafarra, Y.~Hu, P.~Ramadoss, F.~J.~A. Chavez, S.~Traversaro,
  and D.~Pucci, ``A benchmarking of dcm-based architectures for position,
  velocity and torque-controlled humanoid robots,'' \emph{International Journal
  of Humanoid Robotics}, vol.~17, no.~01, p. 1950034, 2020.

\bibitem{Hopkins2015b}
M.~A. Hopkins, D.~W. Hong, and A.~Leonessa, ``{Compliant locomotion using
  whole-body control and Divergent Component of Motion tracking},'' in
  \emph{Proc. - IEEE Int. Conf. Robot. Autom.}, 2015.

\bibitem{Herzog2016}
A.~Herzog, N.~Rotella, S.~Mason, F.~Grimminger, S.~Schaal, and L.~Righetti,
  ``{Momentum control with hierarchical inverse dynamics on a torque-controlled
  humanoid},'' \emph{Auton. Robots}, 2016.

\bibitem{Gilardi2002}
G.~Gilardi and I.~Sharf, ``{Literature survey of contact dynamics modelling},''
  \emph{Mech. Mach. Theory}, 2002.

\bibitem{Whittaker1988}
E.~T. Whittaker and S.~W. McCrae, \emph{{A Treatise on the Analytical Dynamics
  of Particles and Rigid Bodies}}, 1988.

\bibitem{Englsberger2018}
J.~Englsberger, G.~Mesesan, A.~Werner, and C.~Ott, ``{Torque-Based Dynamic
  Walking - A Long Way from Simulation to Experiment},'' 2018.

\bibitem{Mason1988a}
M.~T. Mason and Y.~Wang, ``{On the inconsistency of rigid-body frictional
  planar mechanics},'' in \emph{Proceedings. 1988 IEEE Int. Conf. Robot.
  Autom.}\hskip 1em plus 0.5em minus 0.4em\relax IEEE Comput. Soc. Press, 1988,
  pp. 524--528.

\bibitem{Stronge1991}
W.~J. Stronge, ``{Unraveling paradoxical theories for rigid body collisions},''
  \emph{J. Appl. Mech. Trans. ASME}, 1991.

\bibitem{popov2019handbook}
V.~L. Popov, M.~He{\ss}, and E.~Willert, \emph{Handbook of contact
  mechanics}.\hskip 1em plus 0.5em minus 0.4em\relax Springer Nature, 2019.

\bibitem{Lankarani1990}
H.~M. Lankarani and P.~E. Nikravesh, ``{A contact force model with hysteresis
  damping for impact analysis of multibody systems},'' \emph{J. Mech. Des.
  Trans. ASME}, vol. 112, no.~3, pp. 369--376, sep 1990.

\bibitem{Azad2014}
M.~Azad and R.~Featherstone, ``{A new nonlinear model of contact normal
  force},'' \emph{IEEE Trans. Robot.}, 2014.

\bibitem{Azad2016}
M.~Azad, V.~Ortenzi, H.~C. Lin, E.~Rueckert, and M.~Mistry, ``{Model estimation
  and control of compliant contact normal force},'' in \emph{IEEE-RAS Int.
  Conf. Humanoid Robot.}, 2016.

\bibitem{Henze2016}
B.~Henze, M.~A. Roa, and C.~Ott, ``{Passivity-based whole-body balancing for
  torque-controlled humanoid robots in multi-contact scenarios},'' \emph{Int.
  J. Rob. Res.}, 2016.

\bibitem{Mesesan2019}
G.~Mesesan, J.~Englsberger, G.~Garofalo, C.~Ott, and A.~Albu-Schaffer,
  ``{Dynamic Walking on Compliant and Uneven Terrain using DCM and
  Passivity-based Whole-body Control},'' in \emph{IEEE-RAS Int. Conf. Humanoid
  Robot.}, 2019.

\bibitem{10.1115/1.3139652}
M.~H. Raibert and J.~J. Craig, ``{Hybrid Position/Force Control of
  Manipulators},'' \emph{Journal of Dynamic Systems, Measurement, and Control},
  vol. 103, no.~2, pp. 126--133, 06 1981.

\bibitem{Flayols2020}
T.~Flayols, A.~Prete, M.~Khadiv, N.~Mansard, T.~Flayols, A.~Prete, M.~Khadiv,
  N.~Mansard, L.~R. Balancing, T.~Flayols, A.~D. Prete, M.~Khadiv, N.~Mansard,
  and L.~Righetti, ``{Balancing Legged Robots on Visco-Elastic Contacts},''
  Tech. Rep., 2020.

\bibitem{Catalano2020}
M.~G. Catalano, I.~Frizza, C.~Morandi, G.~Grioli, K.~Ayusawa, T.~Ito, and
  G.~Venture, ``{HRP-4 walks on Soft Feet},'' \emph{IEEE Robot. Autom. Lett.},
  pp. 1--1, mar 2020.

\bibitem{280780}
S.~{Chiaverini}, B.~{Siciliano}, and L.~{Villani}, ``Force/position regulation
  of compliant robot manipulators,'' \emph{IEEE Transactions on Automatic
  Control}, vol.~39, no.~3, pp. 647--652, 1994.

\bibitem{fahmi2019stance}
S.~Fahmi, M.~Focchi, A.~Radulescu, G.~Fink, V.~Barasuol, and C.~Semini,
  ``Stance: Locomotion adaptation over soft terrain,'' 2019.

\bibitem{Li2019}
Q.~Li, Z.~Yu, X.~Chen, Q.~Zhou, W.~Zhang, L.~Meng, and Q.~Huang, ``{Contact
  Force/Torque Control Based on Viscoelastic Model for Stable Bipedal Walking
  on Indefinite Uneven Terrain},'' \emph{IEEE Trans. Autom. Sci. Eng.},
  vol.~16, no.~4, pp. 1627--1639, oct 2019.

\bibitem{doi:10.1177/1729881419897472}
F.~Sygulla and D.~Rixen, ``A force-control scheme for biped robots to walk over
  uneven terrain including partial footholds,'' \emph{International Journal of
  Advanced Robotic Systems}, vol.~17, no.~1, p. 1729881419897472, 2020.

\bibitem{6100848}
K.~{Bouyarmane} and A.~{Kheddar}, ``Fem-based static posture planning for a
  humanoid robot on deformable contact support,'' in \emph{2011 11th IEEE-RAS
  International Conference on Humanoid Robots}, 2011, pp. 487--492.

\bibitem{Natale2017}
L.~Natale, C.~Bartolozzi, D.~Pucci, A.~Wykowska, and G.~Metta, ``{iCub: The
  not-yet-finished story of building a robot child},'' \emph{Sci. Robot.},
  2017.

\bibitem{Marsden2010}
J.~E. Marsden and T.~S. Ratiu, \emph{{Introduction to Mechanics and Symmetry: A
  Basic Exposition of Classical Mechanical Systems}}.\hskip 1em plus 0.5em
  minus 0.4em\relax Springer Publishing Company, Incorporated, 2010.

\bibitem{Olfati-Saber:2001:NCU:935467}
R.~Olfati-Saber, ``{Nonlinear Control of Underactuated Mechanical Systems with
  Application to Robotics and Aerospace Vehicles},'' Ph.D. dissertation,
  Cambridge, MA, USA, 2001.

\bibitem{ljung1999system}
L.~Ljung, \emph{System Identification: Theory for the User}, ser. Prentice Hall
  information and system sciences series.\hskip 1em plus 0.5em minus
  0.4em\relax Prentice Hall PTR, 1999.

\bibitem{Metta2010}
G.~Metta, L.~Natale, F.~Nori, G.~Sandini, D.~Vernon, L.~Fadiga, C.~von Hofsten,
  K.~Rosander, M.~Lopes, J.~Santos-Victor, A.~Bernardino, and L.~Montesano,
  ``{The iCub humanoid robot: An open-systems platform for research in
  cognitive development},'' \emph{Neural Networks}, 2010.

\bibitem{osqp}
B.~Stellato, G.~Banjac, P.~Goulart, A.~Bemporad, and S.~Boyd, ``{OSQP}: an
  operator splitting solver for quadratic programs,'' \emph{Mathematical
  Programming Computation}, vol.~12, no.~4, pp. 637--672, 2020.

\bibitem{4054664}
G.~W.~K. {Colman} and J.~W. {Wells}, ``On the use of rls with covariance reset
  in tracking scenarios with discontinuities,'' in \emph{2006 Canadian
  Conference on Electrical and Computer Engineering}, 2006, pp. 693--696.

\end{thebibliography}
\bibliographystyle{IEEEtran}

\end{document}